\documentclass{article}
\PassOptionsToPackage{numbers, compress}{natbib}

\usepackage{xspace}
\usepackage{amssymb}
\usepackage{amsmath}
\usepackage{multirow}
\usepackage[dvipsnames]{xcolor}
\usepackage{colortbl} % For cell and row coloring
\usepackage{lineno}   % For adding line numbers (optional)
\definecolor{cvprblue}{rgb}{0.21,0.49,0.74}

\usepackage[pagebackref,breaklinks,colorlinks,allcolors=cvprblue]{hyperref}
\usepackage{enumitem}

\usepackage[preprint]{neurips_2025}

\usepackage{textcomp} 
\usepackage{caption}
\usepackage[utf8]{inputenc} % allow utf-8 input
\usepackage[T1]{fontenc}    % use 8-bit T1 fonts
\usepackage{hyperref}       % hyperlinks
\usepackage{url}            % simple URL typesetting
\usepackage{booktabs}       % professional-quality tables
\usepackage{amsfonts}       % blackboard math symbols
\usepackage{nicefrac}       % compact symbols for 1/2, etc.
\usepackage{microtype}      % microtypography
\usepackage{xcolor}         % colors
\usepackage{graphicx}
\usepackage{makecell} % optional, for better control
\usepackage{ulem} % For strikethrough
\usepackage{wrapfig}
\usepackage{floatflt}
\newcommand{\worldwideweb}{\raisebox{-1.5pt}{\includegraphics[height=1.05em]{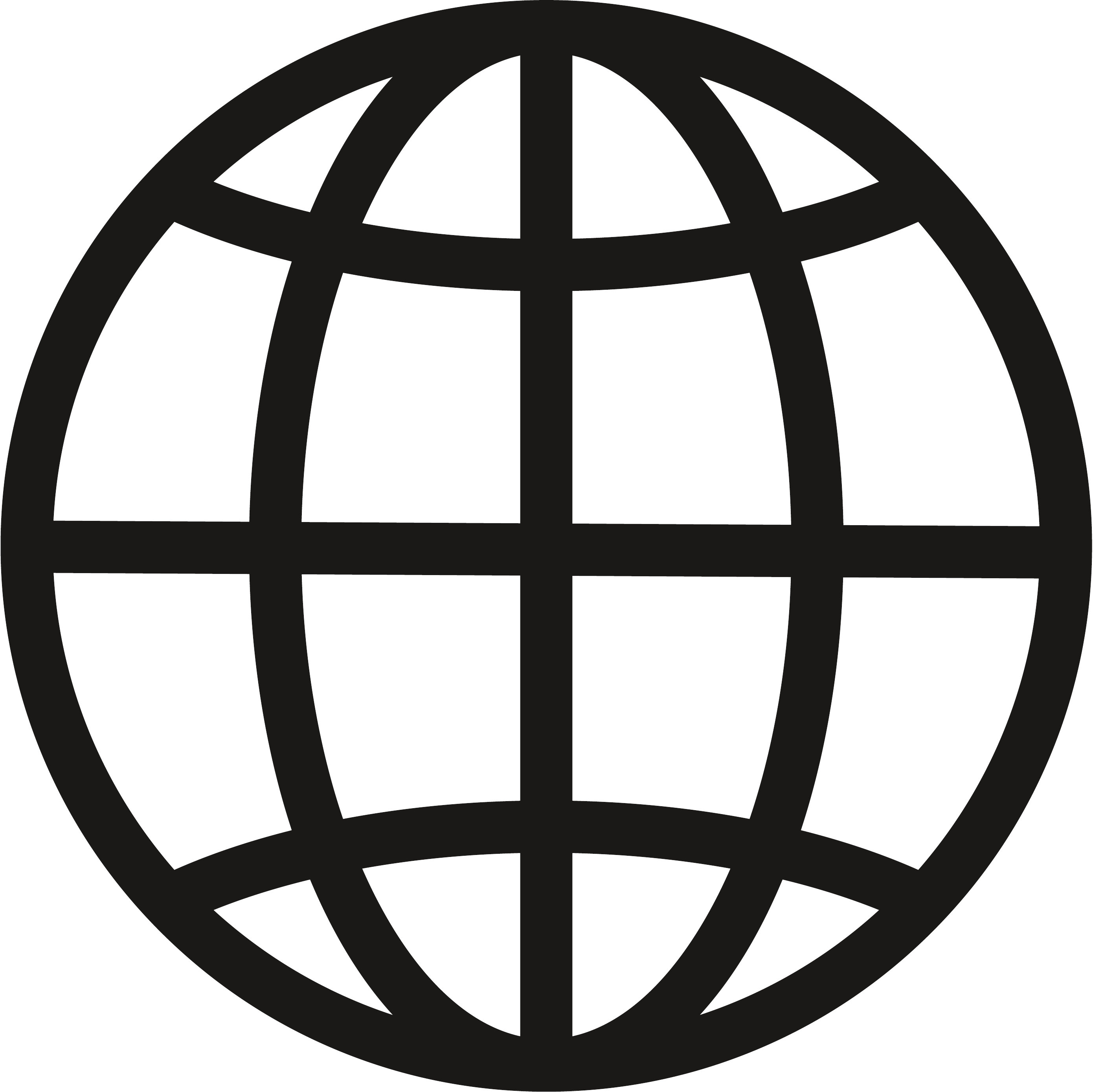}}\xspace}
\newcommand{\github}{\raisebox{-1.5pt}{\includegraphics[height=1.05em]{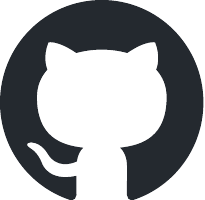}}\xspace}
\newcommand{\huggingface}{\raisebox{-1.5pt}{\includegraphics[height=1.05em]{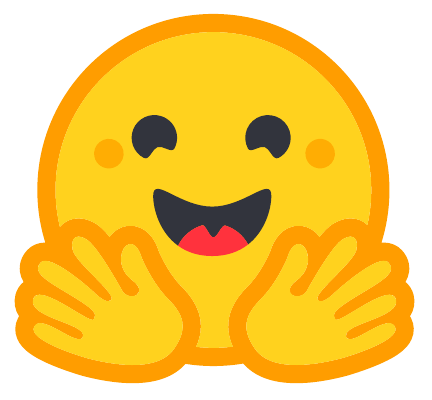}}\xspace}
\title{Spatial Reasoning with Vision-Language Models in Ego-Centric Multi-View Scenes}

\author{%
  Mohsen Gholami\textsuperscript{1}\thanks{Corresponding Author: mohsen.gholami1@huawei.com} , Ahmad Rezaei\textsuperscript{1} , Zhou Weimin\textsuperscript{2} , Sitong Mao\textsuperscript{2} , Shunbo Zhou\textsuperscript{2} , \\ \textbf{Yong Zhang\textsuperscript{1}}\thanks{Equal Contribution}  , \textbf{ Mohammad Akbari\textsuperscript{1}\footnotemark[2]\addtocounter{footnote}{-1}\footnotemark} \\
  \textsuperscript{1} Huawei Technologies Canada, \textsuperscript{2} Huawei Cloud \\
  {\worldwideweb \href{https://vbdi.github.io/Ego3D-Bench-webpage/}{{\text{ Project Page}}}} \quad \quad {\github \href{https://github.com/vbdi/Ego3D-Bench}{{\text{ Code}}}}
  \quad \quad
  {\huggingface \href{https://huggingface.co/datasets/vbdai/Ego3D-Bench}{{\text{ Ego3D-Bench}}}}
  \vspace{0.2cm}
}

\begin{document}

\maketitle
\vspace{-22pt}
\begin{figure}[h!]
    \centering
    \includegraphics[width=1.0\linewidth]{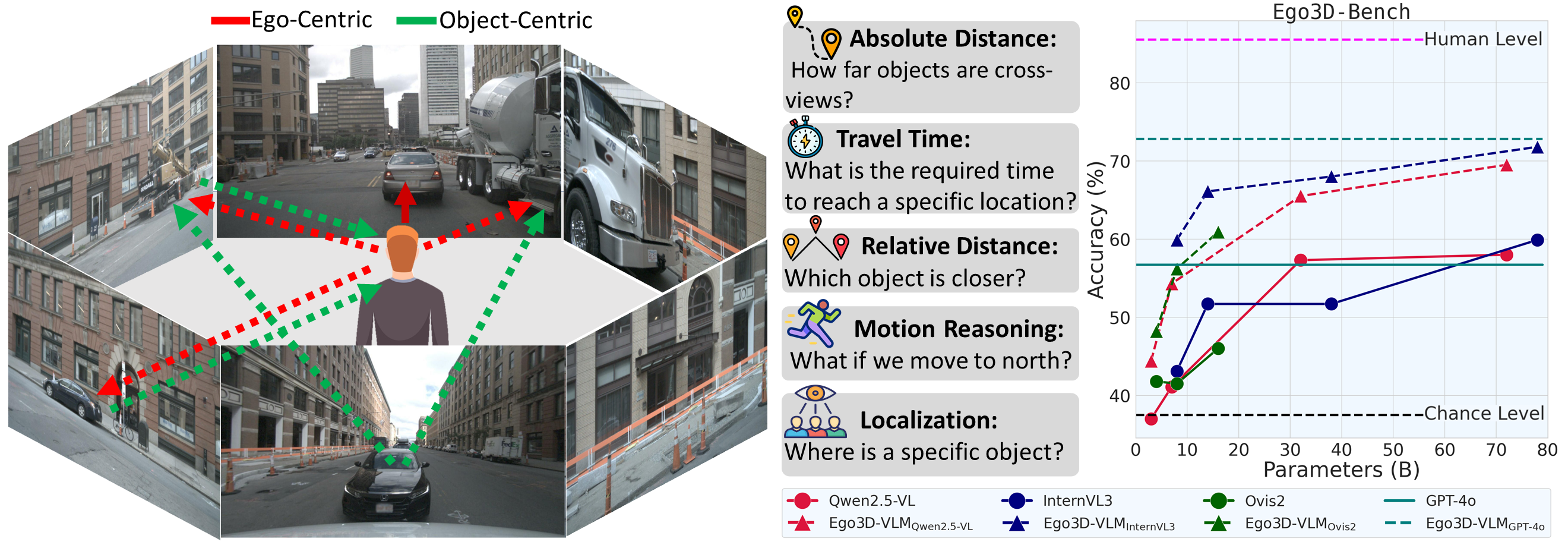}
    \label{fig:enter-label}
    \vspace{-10pt}
    \caption{\texttt{Ego3D-Bench} is {a} {3D} spatial benchmark for VLMs using ego-centric multi-view images. {It} spans ego- and object-centric perspectives across 5 categories. A significant gap exists between human and VLM performance; our method, \textit{Ego3D-VLM}, consistently narrows this gap. 
    }
\end{figure}

\begin{abstract}
\vspace{-10pt}
Understanding 3D spatial relationships remains a major limitation of current Vision-Language Models (VLMs). Prior work has addressed this issue by creating spatial question-answering (QA) datasets based on single images or indoor videos. However, real-world embodied {AI} agents—such as robots and self-driving cars—typically rely on ego-centric, multi-view observations. To this end, we introduce \texttt{Ego3D-Bench}, a new benchmark designed to evaluate the spatial reasoning abilities of VLMs using ego-centric, multi-view outdoor data. \texttt{Ego3D-Bench} comprises over \textbf{8,600} QA pairs, created with significant involvement from human annotators to ensure quality and diversity. We benchmark 16 SOTA VLMs, including GPT-4o, Gemini1.5-Pro, InternVL3, and Qwen2.5-VL. Our results reveal a notable performance gap between human level scores and VLM performance, highlighting that current VLMs still fall short of human level spatial understanding. 
To bridge this gap, we propose \textit{Ego3D-VLM}, a post-training framework that enhances {3D} spatial reasoning of VLMs. \textit{Ego3D-VLM} generates cognitive map based on estimated global 3D coordinates, resulting in \textbf{12\%} and \textbf{56\%} average improvements on multi-choice QA and absolute distance estimation, respectively. \textit{Ego3D-VLM} can be integrated with any existing VLM. Together, \texttt{Ego3D-Bench} and \textit{Ego3D-VLM} offer valuable tools for advancing toward human level spatial understanding in real-world, multi-view environments.
\end{abstract}

\vspace{-10pt}
\section{Introduction}
\vspace{-5pt}
3D spatial understanding  is a critical capability for embodied AI agents operating in the real world \citep{chaplot2021seal, wang2024embodiedscan}. This includes perceiving the location of surrounding objects, estimating their distances, and reasoning about their motion. VLMs have recently emerged as powerful tools to integrate visual perception and language reasoning, making them promising components for building intelligent embodied AI systems \citep{ma2024survey, LLaVA-NeXT-Interleave}. Thus, enhancing and evaluating the spatial understanding of VLMs has become an increasingly important research direction \citep{spatiallm,spatialrgpt,spatialvlm,zhao2025urbanvideobench,zhang2025do, wang2024picture, wu2024mind, RG-SAN, robospatial, SAT}. 

Recent benchmarks on spatial understanding focus mainly on spatial reasoning from single images \citep{spatialrgpt, spatialvlm, QSpatial-Bench} or videos captured in static indoor environments \citep{vis-bench, mm-spatial}. In such cases, spatial understanding is framed around passive observations, where a single camera moves in a room to create a video, and the model should infer spatial relationships in a relatively static scene. This setup differs fundamentally from the perceptual experience of real-world embodied agents such as self-driving cars or mobile robots. These agents rely on ego-centric multi-view observations \citep{ma2024survey}, provided by multiple cameras simultaneously capturing front, side, and rear views of their surroundings. These views are not interchangeable or purely visual; they carry explicit spatial semantics tied to the agent’s frame of reference. E.g, “left” \& “right” refer to fixed directions relative to the agent’s body, and must be interpreted consistently over time as the agent moves through dynamic environments. This distinction is crucial: while prior video-based datasets may offer multiple viewpoints, they do not reflect the structured, directional, and temporally evolving nature of ego-centric multi-view inputs. Also, existing benchmarks do not evaluate VLMs' reasoning ability across these spatially grounded perspectives in dynamic, real-world scenes.

This gap motivates the need for new benchmarks that better align with the spatial reasoning demands of embodied agents. 
To this end, we introduce \texttt{Ego3D-Bench}, a benchmark of 8.6K QA pairs carefully curated from the validation set of 3 public datasets: NuScenes \citep{nuscenes}, Waymo Open Dataset \citep{waymo}, and Argoverse 1 \citep{Argoverse}. Human annotators played a central role in both the dataset construction and the rigorous quality review process to ensure the reliability of the benchmark. We focused specifically on ego-centric multi-view tasks, rather than building a general-purpose benchmark, to complement existing monocular spatial benchmarks (e.g., VSI-Bench \citep{vis-bench}). Thus, we excluded questions that can be answered based on each view independently across multiple images (e.g., counting objects in each image) or general knowledge of LLMs (e.g., estimating the size of well-known objects). To our knowledge, \texttt{Ego3D-Bench} is the first benchmark to evaluate spatial reasoning of VLMs given ego-centric multi-view inputs.

We evaluated {16} SOTA VLMs, including generalist and 3D spatial ones on \texttt{Ego3D-Bench}, revealing a significant gap between human performance and current VLMs. We hypothesize that a key limitation lies in the inability of VLMs to construct a coherent world model from multi-view images. In contrast, humans naturally integrate visual info from their left, right, and front views into a unified spatial representation, enabling real-time reasoning and navigation. Prior work has attempted to bridge this gap by first generating a 3D point-cloud \citep{spatiallm, leo, 3dllm, ChatScene, 3D-CLR} or rendering a bird-eye-view (BEV) image of the scene \citep{GPT4Scene}. Although point-clouds and BEV images offer rich spatial information, they are challenging to reconstruct in dynamic environments, struggle with sparse multi-view inputs, and significantly increase inference time—often by a factor of ten \citep{LLaVA-3D}.

To this end, we propose, \textit{Ego3D-VLM}, a post-training method that improves 3D spatial understanding of VLMs. The main idea of \textit{Ego3D-VLM} is to create a \textbf{textual cognitive map} of the surrounding. The textual cognitive map defines a coordinate system center on the ego and locates important object (i.e., those referred to in the input prompt) in 3D coordinate space. Unlike point-clouds and BEV image methods \citep{spatiallm,GPT4Scene}, our cognitive map only focuses on referred objects, making the number of input tokens significantly smaller and enabling efficient reasoning.

Given multi-view images as input, we first use referring expression comprehension (REC) models to find the 2D location of referred expressions in pixel space. We also use a metric depth estimator to estimate the depth values. We then convert the 2D points to 3D points in camera coordinate space and transform 3D points from all views to the global coordinate space (i.e., front camera coordinate space). We define a cognitive map generator function that returns a textual cognitive map given 3D coordinates. The textual cognitive map defines a coordinate space for the VLM and organizes detected objects (expressions) based on the view-point. The textual cognitive map consistently improves SOTA VLMs on spatial reasoning. The major contributions of our work are: 
\vspace{-5pt}
\begin{itemize}[leftmargin=*]
    \item We introduce \texttt{Ego3D-Bench}, an ego-centric multi-view benchmark to evaluate 3D spatial understanding of VLMs.
    \item We propose \textit{Ego3D-VLM}, a plug-and-play post-training framework to enhance 3D spatial understanding of VLMs, especially in ego-centric multi-view scenarios. 
    \item Through extensive experiments, we demonstrate that \textit{Ego3D-VLM} significantly improves SOTA generalist as well as 3D spatial VLMs in 3D reasoning.
\end{itemize}

\vspace{-5pt}
\section{Related Work}
\vspace{-5pt}
\textbf{3D Spatial Benchmarks/Datasets for VLMs.} 
% VSI-Bench \citep{vis-bench}, CA-VQA \citep{mm-spatial}, Q-Spatial-Bench \citep{QSpatial-Bench}, SpatialRGPT-Bench \citep{spatialrgpt}, RoboSpatial \cite{robospatial}, SAT \cite{SAT}, and All-Angle Bench \citep{all-angle} are recent spatial benchmarks/datasets for VLMs. 
% We categories prior datasets and benchmarks into three categories depending on the input data type. 
% (1) Q-Spatial-Bench \citep{QSpatial-Bench}, RoboSpatial  \citep{robospatial}, and SpatialRGPT-Bench \citep{spatialrgpt}  focus on single-view images. 
% (2) VSI-Bench \citep{vis-bench}, CA-VQA \citep{mm-spatial}, and SAT \citep{SAT} focus on video captured from indoor static scenes. 
% As noted, this setup differs from the perceptual experience of real-world embodied agents, i.e., ego-centric multi-view images. (3) All-Angle Bench \citep{all-angle} is the 1st multi-view benchmark for VLMs. However, in their setup multiple cameras look at a scene from different direction---i.e., a setup for surveillance cameras and motion capture systems. \texttt{Ego3D-Bench} is the first ego-centric multi-view benchmark for VLMs created from dynamic outdoor scenes.
We categorize prior datasets and benchmarks into three groups based on the input data type.
(1) \citep{visualspatial, whatsup, QSpatial-Bench, robospatial, zhang2025do, spatialrgpt,du-etal-2024-embspatial} focus on single-view images.
(2) \citep{vis-bench, mm-spatial, ViewSpatial-Bench, yang2025mmsi, SAT} focus on videos captured from indoor static scenes. As noted, this setup differs from the perceptual experience of real-world embodied agents, which involves ego-centric multi-view images.
(3) All-Angle \cite{all-angle} is the first multi-view benchmark for VLMs. However, in their setup, multiple cameras observe a scene from different directions—similar to surveillance or motion capture systems. In contrast, \texttt{Ego3D-Bench} is the first ego-centric multi-view benchmark for VLMs created from dynamic outdoor scenes.
\begin{figure}[t!]
    \centering
    \vspace{-5pt}
    \includegraphics[width=1.0\linewidth]{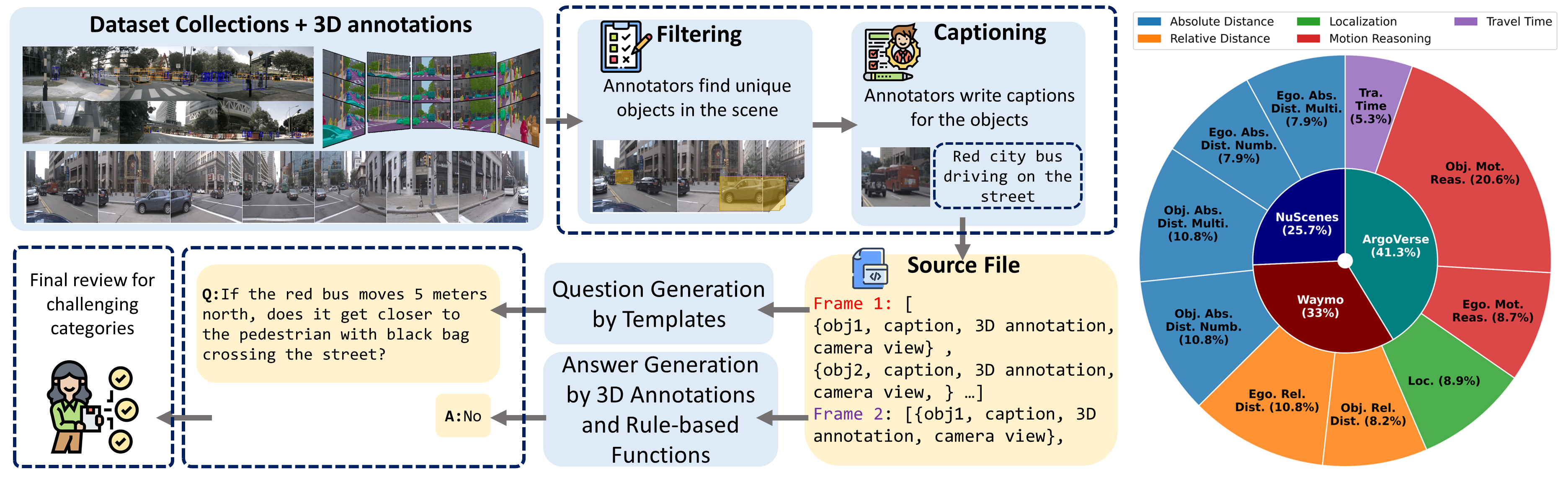}
    \vspace{-3pt}
    \caption{Overview of our \texttt{Ego3D-Bench} creation pipeline. Human annotators played a key role in the process. Right: the distribution of the samples. {\textit{Ego.}: Ego-centric, \textit{Obj.}: Object-centric. 
    }
    }
    \label{fig:benchmark}
    \vspace{-10pt}
\end{figure}

\textbf{3D Spatial VLMs.} 
These models—also referred to as 3D-LLMs or 3D-MLLMs—aim to perform tasks such as 3D grounding, spatial reasoning, depth estimation, and distance measurement. We categorize existing approaches into two main groups: (1) models that take point-clouds as input or reconstruct them from multi-view images, and (2) models that operate directly on image data.

% The first group includes 3D-LLM \citep{3dllm}, 3D-CLR \citep{3D-CLR}, LEO \citep{leo}, SpatialLM \citep{spatiallm}, \cite{zhang2025spartund}, and ChatScene \citep{ChatScene}. 
\citep{3dllm,3D-CLR, leo, spatiallm, zhang2025spartund, ChatScene} fall in the first group.
While point-cloud representations offer rich spatial information, they are difficult to reconstruct in dynamic scenes, often struggle with sparse multi-view data, and significantly increase inference time by over 10×. The 2nd group includes LLaVA-3D \citep{LLaVA-3D}, Video-3D LLM \citep{Video-3D-LLM}, GPT4Scene \citep{GPT4Scene}, MM-Spatial \citep{mm-spatial}, SpatialVLM \citep{spatialvlm}, SpatialRGPT \citep{spatialrgpt}, and SpatialPIN \citep{ma2024spatialpin}. Our work falls in this image-based category.  

LLaVA-3D and Video-3D LLM use depth maps and camera poses for 3D positional encoding. GPT4Scene leverages BEV to address 3D queries. However, these models are primarily trained on indoor, static scenes and limited in handling quantitative spatial relationships such as object distances. SpatialVLM addresses this limitation by introducing a synthetic data generation pipeline, enabling more robust spatial reasoning \citep{VQASynth}. SpatialRGPT enhances input representation using region proposals alongside the original image. MM-Spatial proposes a VLM that supports Chain-of-Thought spatial reasoning involving 2D grounding and depth estimation, and can also leverage depth input via tool-use. Different from SpatialRGPT, SpatialVLM, and MM-Spatial, our proposed \textit{Ego3D-VLM} is a post-training method, can be applied to any existing VLM, and enhances spatial understanding of VLMs and outperforms prior works on ego-centric multi-view reasoning.

\vspace{-5pt}
\section{{Ego3D-Bench}}
\vspace{-5pt}
\texttt{Ego3D-Bench} is designed to quantitatively evaluate 3D spatial understanding of VLMs from multi-view {outdoor} images. A major distinction between \texttt{Ego3D-Bench} and previous works is its focus on ego-centric multi-view data—an approach particularly relevant for applications in self-driving and robotics. 
\texttt{Ego3D-Bench} contains over 8.6K QA pairs in 5 distinct categories. This benchmark is constructed from the validation sets of 3 prominent outdoor multi-view datasets: nuScenes \citep{nuscenes}, Waymo Open Dataset \citep{waymo}, and Argoverse 1 \citep{Argoverse}, featuring 6, 7, and 5 distinct camera views, respectively (Figure \ref{fig:benchmark}).
These datasets cover diverse outdoor environments, including urban areas, highways, and rural regions. \texttt{Ego3D-Bench} leverages the multi-view nature of these datasets to formulate questions that require fine-grained visual comprehension across different viewpoints as well as reasoning over 3D spatial relationships. In this section, we describe the detailed process of constructing the questions of \texttt{Ego3D-Bench}.% (Figure \ref{fig:benchmark}).

\begin{figure}[t!]
    \centering
    \vspace{-5pt}
    \includegraphics[width=1.0\linewidth]{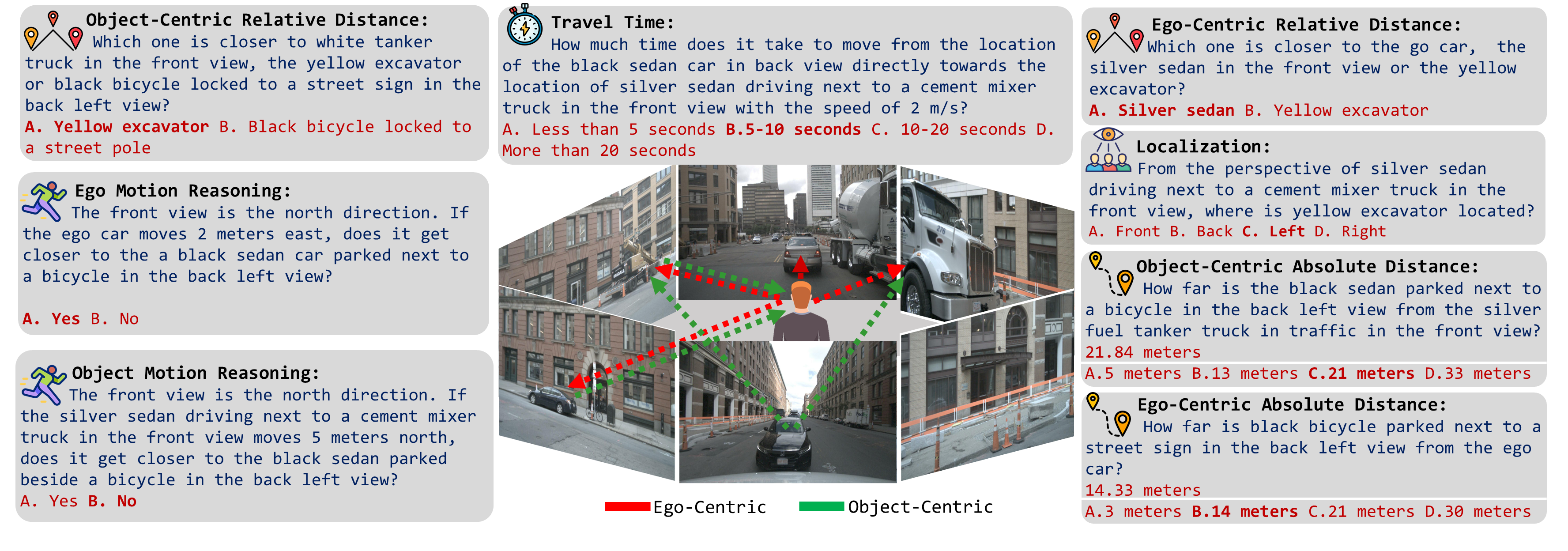}
    \vspace{-15pt}
    \caption{Ego- and object-centric samples from each category in \texttt{Ego3D-Bench}.}
    \label{fig:questions}
    \vspace{-15pt}
\end{figure}

\vspace{-5pt}
\subsection{Benchmark Construction}
\vspace{-5pt}
\textbf{Creation of Source Files:}
{Outdoor datasets, 
%such as the ones used in this work,
unlike their indoor counterparts, often feature multiple instances of the same object within each scene. While indoor environments typically contain unique items—such as a single oven or television per room—outdoor scenes commonly include numerous similar objects, like multiple cars or pedestrians. This makes it challenging to uniquely reference a target object in the scene.}
Thus, annotators begin by carefully reviewing each scene to identify unique objects (called "Filtering" in Figure \ref{fig:benchmark}). {They then compose concise captions describing these objects (called "Captioning" in Figure \ref{fig:benchmark}).} 
The descriptions are designed to be short, yet discriminative such that each object can be uniquely identified. Furthermore, the ground-truth 3D annotations (bounding boxes) are collected from the source datasets.
Object captions, 3D annotations, and the camera view from which the object is visible are used to construct a source file for QA creation. 

\textbf{Creation of QAs:}
{
Each question category follows a predefined template with placeholders for object names, such as \textit{How far is \textless obj1\textgreater in \textless view1\textgreater from \textless obj2\textgreater in \textless view2\textgreater?} (see Appendix \ref{ssec:app_templates} for all question templates). Each question is constructed by placing the generated object annotation and camera views from the source file in the question template. To generate the answers, we use rule-based functions that leverage ground-truth 3D annotations. For challenging categories, such as motion reasoning, a final human review is conducted to ensure the accuracy of the QA pairs.

\vspace{-5pt}
\subsection{Benchmark Details}
\vspace{-5pt}
{Figure \ref{fig:questions} illustrates different question categories in \texttt{Ego3D-Bench}. 
We emphasize on questions that require understanding the 3D world by utilizing information from multiple views. Thus, we exclude questions that can be answered using a single-view or by analyzing each view independently (e.g., counting the number of objects), as they do not contribute to multi-view spatial reasoning.  
We define question from the perspective of both ego and objects in the scene. To clearly indicate the perspective of each question, we categorize them into ego-centric or object-centric.  
In the following, we describe the five tasks (each composed of ego-centric and object-centric types) in our benchmark.}

\textbf{(1) Absolute Distance Measurement.} This category asks the VLM to estimate the absolute metric distance between the ego and another object in the scene or between two different objects from different views. This category is designed in two forms of multi-choice QA and absolute meter.

\textbf{(2) Relative Distance Measurement.} In this task, the VLM is asked to determine which of two objects is closer—either to the ego or to a specific object, designed in the form of two-choice QA.}

\textbf{(3) Localization.} This category is only object-centric and assesses the VLM’s ability to localize objects within a scene. Specifically, the model is asked to infer the location of object-1 from the perspective of object-2 {in the form of multi-choice QA}. 

\textbf{(4) Motion Reasoning.} This category defines a coordinate system using cardinal directions and asks the VLM that if the ego or object-1 moves in a direction, whether or not it gets closer to or farther away from object-2. Answering this {yes/no} question requires mapping the spatial relationship between the objects and how the distances would change if one object moves in a specific direction.

\textbf{(5) Travel Time.} Given a specific motion speed, this category asks ({via multi-choice QA}) to estimate the required time to move from the location of either the ego or object-1 to the location of object-2.

\textbf{Evaluation Metrics.}
We design most of the questions as multi-choice QA and use accuracy as the evaluation metric. We also have two absolute distance estimation for which we use root mean squared error (RMSE) in meters as the evaluation metric.

\begin{figure}[t!]
    \centering
    \vspace{-10pt}
    \includegraphics[width=1.0\linewidth]{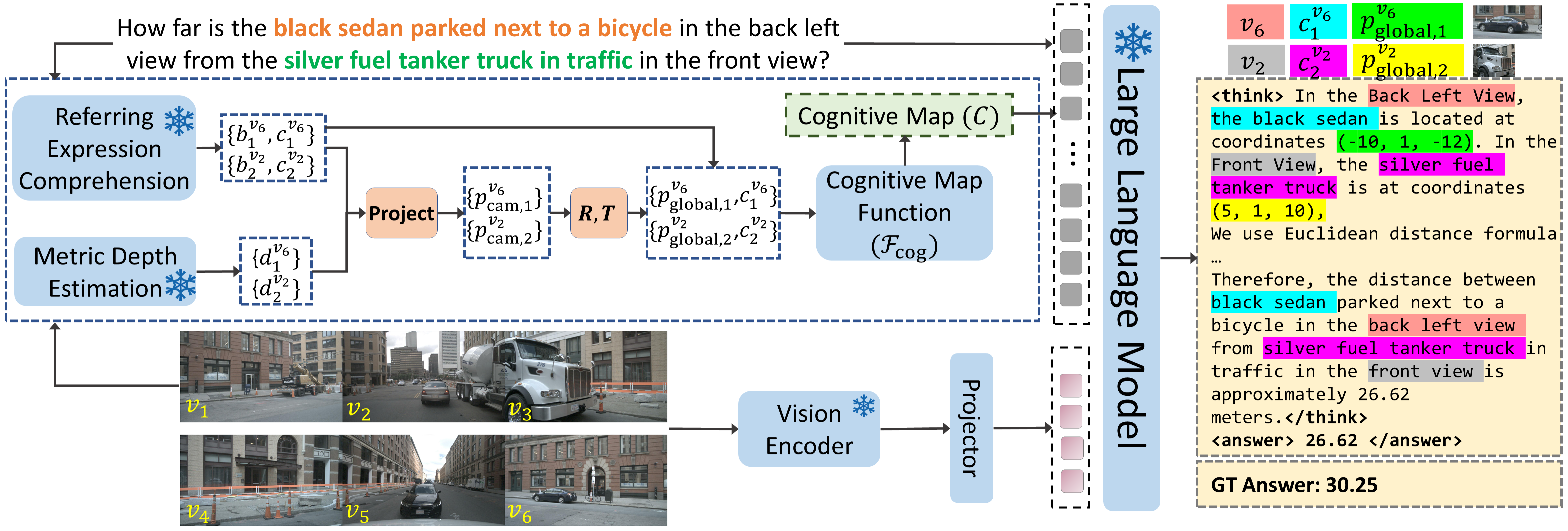}
    \vspace{-10pt}
    \caption{An overview of \textit{Ego3D-VLM}, our post-training 3D spatial understanding framework. 
    }
    \label{fig:framework}
    \vspace{-10pt}
\end{figure}

\vspace{-5pt}
\section{Post-Training 3D spatial understanding : \textit{Ego3D-VLM}}
\vspace{-5pt}
Figure~\ref{fig:framework} shows an overview of our proposed framework, \textit{Ego3D-VLM}. Given a set of multi-view images and a natural language prompt, we use a REC model 
to detect all objects mentioned in the prompt. For each camera view \( v \in \{1, 2, \dots, V\} \), the REC model returns a set of detected objects: 
\begin{equation}
\mathcal{O}^{(v)} = \left\{ \left(b_i^{(v)}, c_i^{(v)}\right) \right\}_{i=1}^{N^{(v)}},     
\end{equation}
where \( b_i^{(v)} \in \mathbb{R}^4 \) denotes the 2D bounding box coordinates for the \(i\)-th object in view \(v\), and \( c_i^{(v)} \) is the corresponding referring expression match. We compute the 2D pixel-space center of each bounding box as $\mathbf{u}_i^{(v)} \in \mathbb{R}^2 $ to create a list of 2D centers of the objects in the pixel space.

\textbf{Camera Coordinate Transformation.} To obtain 3D spatial info, we use a metric depth estimator 
to predict a dense depth map \( D^{(v)} \in \mathbb{R}^{H \times W} \) for each view \(v\). For each detected object center \( \mathbf{u}_i^{(v)} = (x_i, y_i)^\top \), we extract the corresponding depth value \( d_i^{(v)} = D^{(v)}(x_i, y_i) \).
We then project each center point into the 3D camera coordinate system using the camera intrinsics \( K^{(v)} \in \mathbb{R}^{3 \times 3} \):
\begin{equation}
\mathbf{p}_{\text{cam},i}^{(v)} = d_i^{(v)} \cdot \left(K^{(v)}\right)^{-1} \begin{bmatrix} x_i \\ y_i \\ 1 \end{bmatrix} \in \mathbb{R}^3.    
\end{equation}
Next, we transform the 3D point from the local to the global coordinate system, i.e., defined as the front camera view point coordinate system. This replicates the human perception mechanism given multi-view cameras by using the front view as the reference building the 3D world based on that. Using the rotation matrix \( R^{(v)} \in \mathbb{R}^{3 \times 3} \) and translation vector \(T^{(v)} \in \mathbb{R}^{3 \times 3} \), we have:
\begin{equation}
\mathbf{p}_{\text{global},i}^{(v)} = \begin{bmatrix} R^{(v)} & T^{(v)} \\ 0 & 1 \end{bmatrix} \cdot 
\begin{bmatrix}
\mathbf{p}_{\text{cam},i}^{(v)} \\
1
\end{bmatrix}
\in \mathbb{R}^4.
\end{equation}

This process effectively simulates human spatial perception by leveraging multi-view images to construct a unified 3D representation of the scene.

\textbf{Relational Scaling.} Humans estimate object sizes using known references—e.g., knowing a person is $\approx$ 1.7 m tall helps infer the size of nearby objects. Inspired by this, we scale 3D points $\mathbf{p}_{\text{global},i}^{(v)}$ using familiar object categories (e.g., sedans, humans, bikes) identified in a few representative frames across all camera views.
We compute the average observed height $h_{\text{est}}$ and scale all 3D points by $s = {h_{\text{cs}}}/{h_{\text{est}}}$, where $h_{\text{cs}}$ is the canonical common sense height (e.g., 1.7 m for humans). This yields $\mathbf{p}_{\text{scaled},i}^{(v)} = s \cdot \mathbf{p}_{\text{global},i}^{(v)}$, producing physically plausible scales without ground-truth depth.

\begin{figure}[t!]
    \centering
    \vspace{-3pt}
    \includegraphics[width=1.0\linewidth]{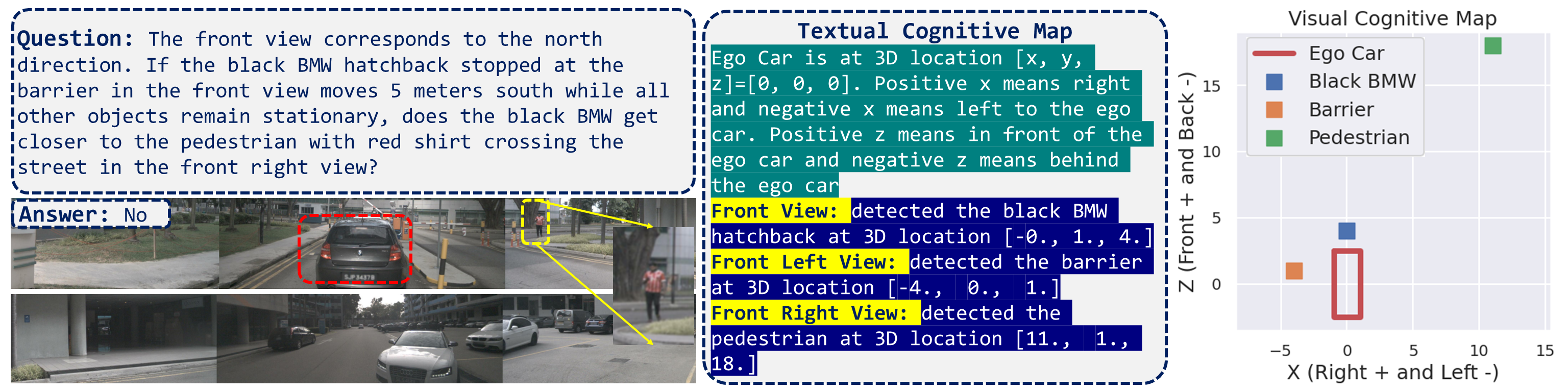}
    \vspace{-10pt}
    \caption{Example question with associated textual and visual cognitive maps. 
    }
    \label{fig:cog_map}
    \vspace{-10pt}
\end{figure}

\textbf{Creating a Cognitive Map.}
We define a cognitive map generator function \(\mathcal{F}_{\text{cog}}\), which takes as input the set of 3D global coordinates and corresponding referring expressions for all detected objects across all camera views. Specifically, for each object \(i\) detected from view \(v\), we denote its global 3D position as \(\mathbf{p}_{\text{global},i}^{(v)}\) and its matched referring expression as \(c_i^{(v)}\). The function outputs a textual cognitive map \(C\), defined as:
\vspace{-5pt}
\begin{equation}
  C = \mathcal{F}_{\text{cog}} \left( \left\{ \left( \mathbf{p}_{\text{global},i}^{(v)}, c_i^{(v)} \right) \right\}_{i,v} \right).  
\end{equation}
\(\mathcal{F}_{\text{cog}}\) constructs an ego-centric world model centered on the agent. It integrates multi-view detections and links each referred object to its spatial position and originating viewpoint. The resulting cognitive map captures both linguistic references and spatial semantics in a compact, human-interpretable form—enabling grounded reasoning and situational awareness. Figure~\ref{fig:cog_map} illustrates a sample prompt, the corresponding multi-view images, and the generated cognitive maps.

Given a VLM \(\mathcal{V}\) that answers queries by visual-textual contexts, it takes as input the cognitive map \(C\), a set of multi-view {images} \(\mathcal{I} = \{I^{(v)}\}_v\), and a natural language query \(q\), to return an answer \(a\):
\begin{equation}
  a = \mathcal{V}(C, \mathcal{I}, q).  
\end{equation}
The cognitive map \(C\) provides structured spatial grounding, while \(\mathcal{I}\) supplies rich visual cues—such as appearance, color, and fine-grained context—not captured in \(C\). Together, they guide the VLM in interpreting and answering the query.

\vspace{-5pt}
\section{Evaluation on \texttt{Ego3D-Bench}}
\vspace{-5pt}
In this section, we present a comprehensive evaluation of VLMs on \texttt{Ego3D-Bench}. We organize our experiments into 4 parts by benchmarking (1) generalist VLMs (i.e., models trained for general vision-language tasks), (2) 3D-VLMs (i.e., models trained specifically for 3D SU), (3) VLM\textsubscript{+Depth+REC} (i.e., generalist VLMs augmented with depth and REC tools), and (4) ablation studies. We use a fixed {R1-style prompt} \citep{guo2025deepseek} in the evaluations of all models (details in the Appendix \ref{ssec:app_prompts}). {We use Grounding-Dino-Base \citep{groundingdino2023hf} as the REC model and Depth-Anything-V2-Metric-Large \citep{depthanythingv2hf} as the metric depth estimator in all experiments.} 

\setlength{\tabcolsep}{3pt} 
\renewcommand{\arraystretch}{0.5}
\begin{table*}[t!]
    \centering
    \footnotesize
    \vspace{-13pt}
    \caption{     
    {Comparison results of generalist VLMs vs. \textit{Ego3D-VLM} on \texttt{Ego3D-Bench}.}
    }   
    \resizebox{1.0\textwidth}{!}{
    \begin{tabular}{l|cccccccc|p{0.65cm}|cc|p{0.65cm}}
          \toprule
          & \multicolumn{9}{c|}{Accuracy (\%) $\uparrow$}&\multicolumn{3}{c}{RMSE$ \downarrow$} \\ \hline \rowcolor[gray]{0.94}
        \rule{0pt}{10pt}& 
         \rotatebox{0}{Ego} &
         \rotatebox{0}{Obj.} &
         \rotatebox{0}{} &
         \rotatebox{0}{Ego} &
         \rotatebox{0}{Obj.} &
         \rotatebox{0}{Travel} &
         \rotatebox{0}{Ego} &
         \rotatebox{0}{Obj.} &
         \rotatebox{0}{} &
         \rotatebox{0}{Ego} &
         \rotatebox{0}{Obj.} &
         \rotatebox{0}{}\\ \rowcolor[gray]{0.94}
         Model&\rotatebox{0}{Dist.} &
         \rotatebox{0}{Dist.} &
         \rotatebox{0}{Loc.} &
         \rotatebox{0}{Mot.} &
         \rotatebox{0}{Mot.} &
         \rotatebox{0}{Time} &
         \rotatebox{0}{Rel.} &
         \rotatebox{0}{Rel.} &
         \rotatebox{0}{\textcolor{blue}{\textbf{Avg.$\uparrow$}}} &
         \rotatebox{0}{Dist.} &
         \rotatebox{0}{Dist.} &
         \rotatebox{0}{\textcolor{blue}{\textbf{Avg.$\downarrow$}}}\\ 
         \midrule
         Human Level &78.2& 57.1&100&100&95.3&72.7&85.0&94.1&\textcolor{blue}{85.3}&-&-&-\\ 
         Chance Level & 25.8 &25.5& 24.3&49.7&50.6&24.0&49.9&50.1&\textcolor{blue}{37.5}& - & - & -\\ 
         \midrule
         \rowcolor[gray]{.90}
         \multicolumn{13}{l}{\textit{Closed-source Models}} \\ % No vertical borders!
         \hspace{2.5mm} GPT-4o &51.4&38.4&47.7&86.9&71.1&35.8&65.9&84.2&\textcolor{blue}{56.7}&8.5&29.5&\textcolor{blue}{19.2}\\
          \hspace{2.5mm} \textcolor{teal}{\textit{\textbf{Ego3D-VLM}} \textsubscript{GPT-4o}}&76.3&58.9&57.3&87.3&89.7&59.9&70.6&85.3 &\textcolor{blue}{\textbf{73.2}}&6.3&8.4&\textcolor{blue}{\textbf{7.4}}\\ \cmidrule(lr){2-13}
         \hspace{2.5mm} Gemini-1.5-Pro & 51.3 &35.8 &49.4&88.4&70.6&37.7&55.5&71.4&\textcolor{blue}{57.5}&10.7&28.6&\textcolor{blue}{19.6}\\
         \hspace{2.5mm} \textcolor{teal}{\textit{\textbf{Ego3D-VLM}} \textsubscript {Gemini-1.5-Pro}} & 65.0&62.0&67.2&93.9&91.4&58.4&66.2&80.6&\textcolor{blue}{\textbf{73.1}}&6.6&7.8&\textcolor{blue}{\textbf{7.2}}\\ 
         \midrule
         \rowcolor[gray]{0.9}
         \multicolumn{13}{l}{\textit{Qwen2.5 Family}} \\
         \hspace{2.5mm} Qwen2.5-3B& 21.5&29.4&28.8&50.3&41.9&30.9&54.1&56.1&\textcolor{blue}{39.1}&30.5&33.7&\textcolor{blue}{32.1}\\ 
         \hspace{2.5mm} \textcolor{teal}{\textit{\textbf{Ego3D-VLM}} \textsubscript {Qwen2.5-3B}} &35.0&30.0&29.8&52.4&56.4&29.6&60.1&62.0&\textcolor{blue}{\textbf{44.4}}&12.7&12.5& \textcolor{blue}{\textbf{12.6}} \\  \cmidrule(lr){2-13}
         \hspace{2.5mm} Qwen2.5-7B&32.7&31.5&30.5&45.9&44.0&34.5&43.2&66.5&\textcolor{blue}{{41.1}}&25.1&35.5&\textcolor{blue}{30.3}\\ 
         \hspace{2.5mm} \textcolor{teal}{\textit{\textbf{Ego3D-VLM}} \textsubscript {Qwen2.5-7B}}&59.4&54.5&33.1&62.3&58.2&49.4&50.5&66.9&\textcolor{blue}{\textbf{54.3}}&8.1&10.9&\textcolor{blue}{\textbf{9.5}} \\ \cmidrule(lr){2-13}
         \hspace{2.5mm} Qwen2.5-32B& 45.4&40.7&49.6&75.6&74.1&40.1&54.0&79.0&\textcolor{blue}{57.3}&21.2&10.4&\textcolor{blue}{15.8}\\
         \hspace{2.5mm} \textcolor{teal}{\textit{\textbf{Ego3D-VLM}} \textsubscript {Qwen2.5-32B}}& 63.7&62.6&54.5&87.7&86.2&40.8&62.6&72.0&\textcolor{blue}{\textbf{65.5}}&11.6&15.9&\textcolor{blue}{\textbf{13.7}}\\ \cmidrule(lr){2-13}
         \hspace{2.5mm} Qwen2.5-72B&42.4&38.6&54.8&86.8&68.9&38.5&53.3&80.5&\textcolor{blue}{58.0}&10.3&22.2&\textcolor{blue}{16.2} \\
         \hspace{2.5mm} \textcolor{teal}{\textit{\textbf{Ego3D-VLM}} \textsubscript {Qwen2.5-72B}}& 62.1&61.4&58.2&94.0&84.5&56.0&63.1&76.6&\textcolor{blue}{\textbf{69.5}}&6.8&8.3&\textcolor{blue}{\textbf{7.5}} 
\\ 
         \midrule
         \rowcolor[gray]{0.9}
         \multicolumn{13}{l}{\textit{InternVL3 Family}} \\
         \hspace{2.5mm} InternVL3-8B&25.8&28.7&29.8&54.1&54.8&36.1&49.9&65.2&\textcolor{blue}{43.1}&15.2&39.1&\textcolor{blue}{27.2} \\
         \hspace{2.5mm} \textcolor{teal}{\textit{\textbf{Ego3D-VLM}} \textsubscript {InternVL3-8B}} &65.4&56.1&37.0
&73.0&71.4&49.0&63.5&66.0&\textcolor{blue}{\textbf{60.1}}&6.8&9.0&\textcolor{blue}{\textbf{8.0}}
\\ \cmidrule(lr){2-13}
         \hspace{2.5mm} InternVL3-14B&46.0&35.6&35.9&63.2&65.9&41.6&55.5&70.1&\textcolor{blue}{51.7}&10.6&24.5&\textcolor{blue}{17.6}\\
         \hspace{2.5mm} \textcolor{teal}{\textit{\textbf{Ego3D-VLM}} \textsubscript {InternVL3-14B}}&70.3&60.9&50.5&79.8&83.1&50.2&63.0&70.7&\textcolor{blue}{\textbf{66.1}}&6.6&8.8&\textcolor{blue}{\textbf{7.7}}\\ \cmidrule(lr){2-13}
         \hspace{2.5mm} InternVL3-38B& 35.4&31.0&39.4&66.6&64.9&38.0&61.0&77.3&\textcolor{blue}{51.7}&8.6&42.2&\textcolor{blue}{25.4} \\
         \hspace{2.5mm} \textcolor{teal}{\textit{\textbf{Ego3D-VLM}} \textsubscript {InternVL3-38B}}& 55.1&64.5&53.4&87.2&88.9&51.9&66.6&76.5& \textcolor{blue}{\textbf{68.0}}&8.2&8.7&\textcolor{blue}{\textbf{8.5}}
\\ \cmidrule(lr){2-13}
         \hspace{2.5mm} InternVL3-78B&54.6&48.4&50.3&77.7&70.0&44.8&57.0&76.6 
        &\textcolor{blue}{59.9}&12.0&15.5&\textcolor{blue}{13.8}
\\ 
         \hspace{2.5mm} \textcolor{teal}{\textit{\textbf{Ego3D-VLM}} \textsubscript{InternVL3-78B}}&68.3&62.7&62.9&91.6&89.2&55.1&66.3&78.2 
&\textcolor{blue}{\textbf{71.8}}&6.8&8.1&\textcolor{blue}{\textbf{7.4}}
\\
         \midrule
         \rowcolor[gray]{0.9}
         \multicolumn{13}{l}{\textit{Ovis2 Family}} \\
         \hspace{2.5mm} Ovis2-4B& 
         29.8&28.9&18.4&47.5&48.1&36.9&54.2&70.3 & \textcolor{blue}{41.8}& 17.1&29.5& \textcolor{blue}{23.3}\\
         \hspace{2.5mm} \textcolor{teal}{\textit{\textbf{Ego3D-VLM}} \textsubscript {Ovis2-4B}} & 62.1&46.9&20.1&49.1&51.9&33.3&62.0&60.4& \textcolor{blue}{\textbf{48.2}}& 6.5& 10.4& \textcolor{blue}{\textbf{8.5}} 
\\ \cmidrule(lr){2-13}
         \hspace{2.5mm} Ovis2-8B&25.6&28.6&31.1&45.3&51.4&31.4&50.7&68.2 & \textcolor{blue}{41.5}& 11.5& 30.2& \textcolor{blue}{20.8}\\
         \hspace{2.5mm} \textcolor{teal}{\textit{\textbf{Ego3D-VLM}} \textsubscript {Ovis2-8B}}&64.9&54.9&33.5&57.2&57.1&46.1&65.1&71.0 &\textcolor{blue}{\textbf{56.2}}&6.0&9.5&\textcolor{blue}{\textbf{7.8}}
\\ 
         \cmidrule(lr){2-13}
         \hspace{2.5mm} Ovis2-16B& 41.7&36.5&41.1&50.5&52.5&27.9&50.9&74.9 & \textcolor{blue}{47.0}& 10.8& 16.6& \textcolor{blue}{13.7}\\
         \hspace{2.5mm} \textcolor{teal}{\textit{\textbf{Ego3D-VLM}} \textsubscript {Ovis2-16B}}&63.4&58.1&43.7&53.9&73.4&51.1&61.0&82.9&\textcolor{blue}{\textbf{60.9}}&6.6&8.3&\textcolor{blue}{\textbf{7.4}}\\
         \bottomrule
    \end{tabular}}
    \vspace{-15pt}
    \label{tab:image-bench}
\end{table*}
% \vspace{-25pt}
\textbf{Chance/Human Performance Levels.}
{We provide frequency-based random selection as the chance level baseline for the multi-choice questions. Furthermore, we conduct human evaluation on multi-choice questions, {where} 10\% of the questions from each category are randomly sampled and evaluated by each human annotator. Table~\ref{tab:image-bench} presents the results of this analysis. While humans can accurately answer the questions that require reasoning about relative location of the objects in space, their performance degrades in questions that require estimation of the exact distance between objects. 
This highlights the challenging nature of accurate distance estimation.}

\vspace{-5pt}
\subsection{Benchmarking Generalist VLMs}
\vspace{-5pt}
We use GPT-4o \citep{openai2024gpt4o} and Gemini-2-Flash \citep{google2023gemini} closed-source VLMs, and {3 competitive families of open-source models: InternVL3 \citep{internvl3}, Qwen2.5-VL \citep{qwen2025qwen25technicalreport}, and Ovis2 \citep{ovis}} open-source models. Inference time analysis, numerical results on more generalist VLMs, and qualitative examples are given in the Appendix \ref{ssec:app_inference_time}, \ref{ssec:app_more_vlms}, and \ref{ssec:app_prompts}.

As given in Table~\ref{tab:image-bench}, the performance of the VLMs varies considerably across different model parameter sizes. Smaller models (e.g., 3B and 8B) operate near chance level, 
indicating limited capacity for multi-view 3D reasoning. In contrast, larger models demonstrate substantial improvements over the chance level, yet still exhibit a noticeable gap when compared to human level performance. Furthermore, our proposed \textit{Ego3D-VLM} 
provides significant improvement across all model sizes and tasks (average 56\% relative improvement in RMSE and 12\% in Accuracy), underscoring the importance of providing structured spatial representations to the model in 3D understanding tasks.

\begin{wrapfigure}{r}{0.38\textwidth}
     \vspace{-7pt} 
    \centering
    \includegraphics[width=1\linewidth]{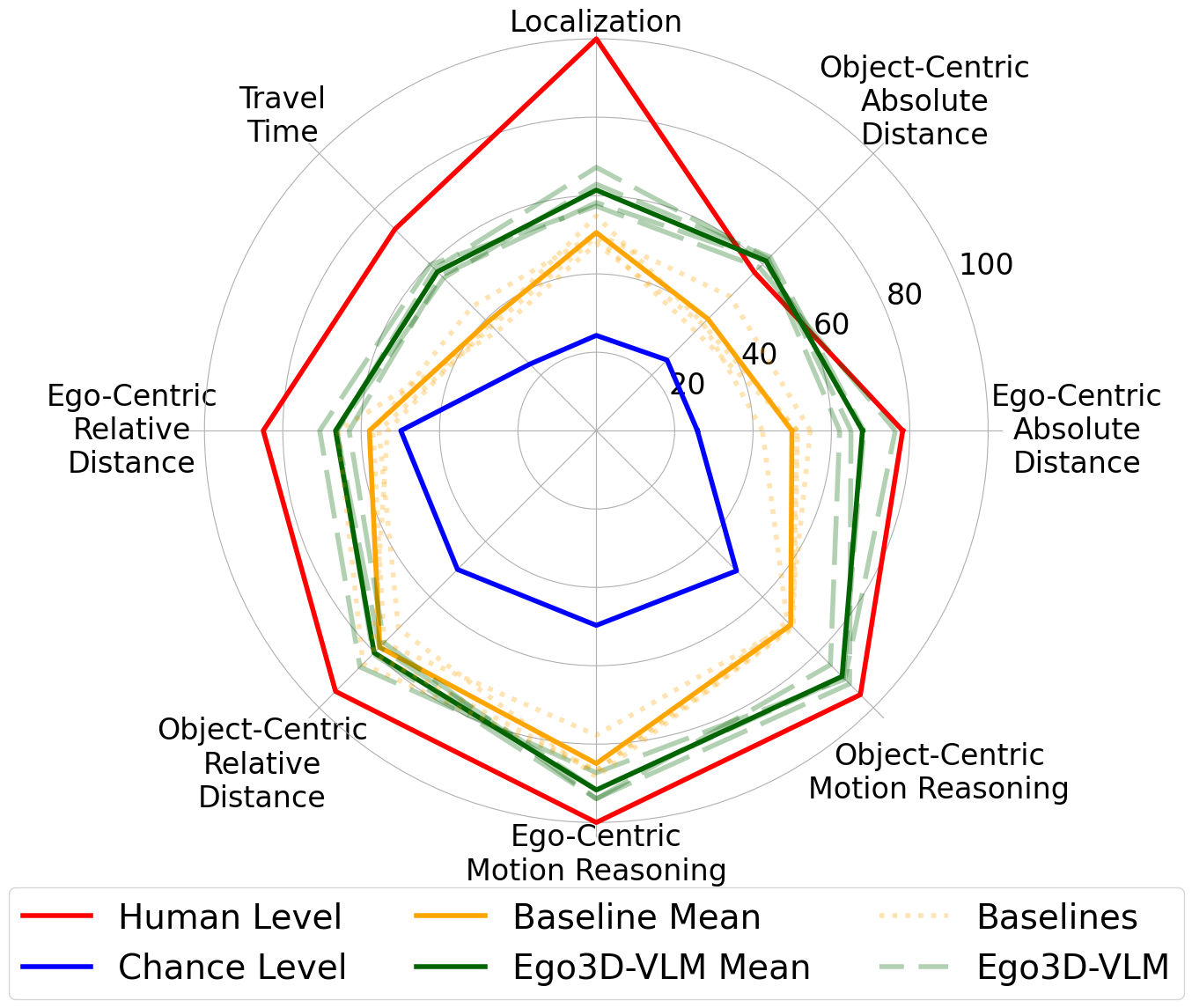}
    \vspace{-10pt}
    %\caption{{Average performance of leading VLMs (GPT-4o, Gemini-1.5-Pro, InternVL3-78B, and Qwen2.5-VL-72B) with/without \textit{Ego3D-VLM} vs. chance and human levels on each category of \texttt{Ego3D-Bench}.}
    \caption{Average performance of leading VLMs w/ and w/o \textit{Ego3D-VLM} vs. chance \& human levels on each category of \texttt{Ego3D-Bench}.}
    \label{fig:mean_task_performance}
    \vspace{-20pt}
\end{wrapfigure}
\textbf{Performance Analysis.} Figure~\ref{fig:mean_task_performance} shows the average performance of 4 leading models {(GPT-4o, Gemini-1.5-Pro, InternVL3-78B, and Qwen2.5-VL-72B)} with and without \textit{Ego3D-VLM} integration across all multiple-choice QAs. \textit{Travel time}, \textit{localization}, and \textit{object-centric absolute distance} are the most challenging tasks for VLMs with 40\%-45\% average accuracy% with the largest models of distinct families
. We argue that these categories require intricate spatial reasoning and the ability to construct an accurate mental map by the relative positioning of objects. While humans can simply build such maps and achieve perfect accuracy, VLMs struggle to replicate this level of SU, indicating a key area for further development. In the absolute distance tasks, the incorporation of 3D location data through the cognitive map substantially narrows the gap between model and human level accuracy. Notably, for object-centric absolute distance, VLMs augmented with \textit{Ego3D-VLM} even surpass human. This is expected, as human estimation of 3D distances in object-centric cases is prone to substantial error without explicit 3D spatial info. Conversely, in the localization task, VLMs continue to fall short of human proficiency, even with cognitive map support. 

\setlength{\tabcolsep}{3pt} 
\renewcommand{\arraystretch}{0.5}
\begin{wraptable}{r}{0.51\textwidth}
\renewcommand{\arraystretch}{0.4}
    \vspace{-12pt}
    \centering
    \footnotesize
    %\caption{Performance of cognitive map fed to LLM.}
    \caption{Blind \textit{Ego3D-VLM} performance analysis.}
    \resizebox{0.50\textwidth}{!}{
    \begin{tabular}{l|c|cc}
        \toprule \rowcolor[gray]{0.94}
         &{Only}&{{Mult. Choice}} &{{Abs. Ans.}}\\ \rowcolor[gray]{0.94}
         Model &{LLM}&(Acc. $\uparrow$)&(RMSE$\downarrow$)\\ \midrule
         InternVL3-8B&&43.1&27.2\\
         Ego3D-VLM\textsubscript{InternVL3-8B}& \checkmark&53.1&11.4 \\
         Ego3D-VLM\textsubscript{InternVL3-8B}& &\textbf{60.1}&\textbf{8.0}\\ \midrule
         InternVL3-14B&&51.7&17.6\\
         Ego3D-VLM\textsubscript{InternVL3-14B} & \checkmark&55.6&10.8 \\
         Ego3D-VLM\textsubscript{InternVL3-14B}& &\textbf{68.0}&\textbf{8.5}\\
         \bottomrule
    \end{tabular}}
    \label{tab:llm}
    % \vspace{6pt}   
\end{wraptable}
\textbf{Blind VLM Performance.} This baseline evaluates how much spatial reasoning can be achieved by VLMs using only textual input, without any visual information, relying solely on their world knowledge \citep{Majumdar_2024_CVPR}. We report the average performance of {GPT-4o and Gemini-1.5-Pro models}. %(2 open-source and 2 closed-source).
The blind VLMs perform {5\%} worse than vision-enabled VLMs ({53.8\%} vs. 58.8\%) and {16.4\%} better than chance level. These results are consistent with findings from prior single-view and video-based spatial reasoning benchmarks \citep{spatialrgpt,vis-bench}.

\textbf{Blind \textit{Ego3D-VLM} Performance.} 
we use the same models for fair comparison. 
Table \ref{tab:llm} shows that Ego3D-VLM with VLM outperforms Ego3D-VLM with LLM since VLMs can ignore false positives in the cognitive map and remain robust when false negatives occur, whereas LLMs suffer performance drops in those cases.

\subsection{Benchmarking 3D-VLMs}
3D-VLM models have been trained on datasets designed for 3D SU, such as absolute distance estimation and relative location inference. We benchmark SpatialRGPT and two checkpoints trained with the Spatial-VLM framework \citep{spatialvlm}: SpaceThinker-Qwen2.5-3B \citep{VQASynth} and SpaceQwen2.5-3B. The SpatialRGPT \citep{spatialrgpt} model assumes that specific regions of the image are annotated with bounding boxes and is trained to answer 3D questions based on these regions. To evaluate SpatialRGPT on our \texttt{Ego3D-Bench}, we reformat the input: object names in the questions are replaced with placeholder labels (e.g., \textit{region-i}), and a list of corresponding region captions is passed to the REC model 
to generate bounding boxes. These estimated bounding boxes were then overlaid on the images before being fed to SpatialRGPT.

Table~\ref{tab:3d-VLMs} presents the performance of the above-mentioned {3D-VLMs on our \texttt{Ego3D-Bench}. SpaceThinker-Qwen2.5-3B achieves the highest overall performance, surpassing both SpatialRGPT with 8B parameters and other generalist VLMs of similar scale. This outcome highlights the critical role of dedicated 3D spatial pretraining and end-to-end architecture in advancing VLM capabilities for spatial reasoning tasks. Moreover, augmenting SpaceThinker with \textit{Ego3D-VLM} leads to an average improvement of 3\% on multiple-choice questions and a reduction of more than 4 meters in absolute distance RMSE, emphasizing the effectiveness of our proposed solution.}

\setlength{\tabcolsep}{3pt} 
\renewcommand{\arraystretch}{0.7}
\begin{table}[h!]
    \centering
    \footnotesize
    \vspace{-12pt}
    \caption{ 
    {Comparison results of 3D-VLMs and our method (\textit{Ego3D-VLM}) on \texttt{Ego3D-Bench}.}  
    }
    \resizebox{0.9\textwidth}{!}{
    \begin{tabular}{l|cccccccc|p{0.65cm}|cc|p{0.65cm}}
          \toprule
          & \multicolumn{9}{c|}{Accuracy (\%) $\uparrow$}&\multicolumn{3}{c}{RMSE$ \downarrow$} \\ \hline \rowcolor[gray]{0.94}
        \rule{0pt}{10pt}& 
         \rotatebox{0}{Ego} &
         \rotatebox{0}{Obj.} &
         \rotatebox{0}{} &
         \rotatebox{0}{Ego} &
         \rotatebox{0}{Obj.} &
         \rotatebox{0}{Travel} &
         \rotatebox{0}{Ego} &
         \rotatebox{0}{Obj.} &
         \rotatebox{0}{} &
         \rotatebox{0}{Ego} &
         \rotatebox{0}{Obj.} &
         \rotatebox{0}{}\\ \rowcolor[gray]{0.94}
         Model&\rotatebox{0}{Dist.} &
         \rotatebox{0}{Dist.} &
         \rotatebox{0}{Loc.} &
         \rotatebox{0}{Mot.} &
         \rotatebox{0}{Mot.} &
         \rotatebox{0}{Time} &
         \rotatebox{0}{Rel.} &
         \rotatebox{0}{Rel.} &
         \rotatebox{0}{\textcolor{blue}{\textbf{Avg.$\uparrow$}}} &
         \rotatebox{0}{Dist.} &
         \rotatebox{0}{Dist.} &
         \rotatebox{0}{\textcolor{blue}{\textbf{Avg.$\downarrow$}}}\\ 
         \midrule 
         SpaceQwen2.5-VL-3B&18.1&21.4&16.1&35.6&35.2&30.2&31.2&32.0&\textcolor{blue}{27.5}&10.6&15.7&\textcolor{blue}{13.2} \\
         SRGPT-VILA1.5-8B&45.7&35.5&39.6&43.6&45.8&24.9&50.8&71.7&\textcolor{blue}{44.7}&11.5&15.1&\textcolor{blue}{13.3} \\ 
         SThinker-Qwen2.5-3B&38.9&39.0&21.9&57.5&53.4&27.1&52.8&71.6&\textcolor{blue}{45.2}& 15.2&16.9&\textcolor{blue}{16.0} \\ \midrule
         \textcolor{teal}{\textit{\textbf{Ego3D-VLM}}\textsubscript{SThinker-Qwen2.5-3B}} & 50.6&44.2&26.4&53.4&57.0&42.6 &54.7&59.8 &\textcolor{blue}{\textbf{48.6}}&11.1&12.2&\textcolor{blue}{\textbf{11.6}}
\\ 
         \bottomrule
    \end{tabular}}
     \vspace{-10pt}    
    \label{tab:3d-VLMs}
\end{table}

%\vspace{-5pt}
\subsection{Benchmarking  VLM\textsubscript{+Depth+REC}.}
\vspace{-5pt}
\setlength{\tabcolsep}{3pt} 
\renewcommand{\arraystretch}{0.5}
\begin{wraptable}{r}{0.45\textwidth}
    \vspace{-13pt}
    \centering
    \footnotesize
    \caption{\textit{Ego3D-VLM} vs. generalist VLMs with REC+depth tools on \texttt{Ego3D-Bench}.}
    \resizebox{0.45\textwidth}{!}{
    \begin{tabular}{l|cc}
        \toprule \rowcolor[gray]{0.94}
         &{{Mult. Choice}} &{{Abs. Ans.}}\\ \rowcolor[gray]{0.94}
         Model &(Acc. $\uparrow$)&(RMSE$\downarrow$)\\ \midrule
         InternVL3-8B&43.1&27.2\\
         InternVL3-8B\textsubscript{+Depth+REC}&51.6&13.1\\
         \textcolor{teal}{\textit{\textbf{Ego3D-VLM}} \textsubscript {InternVL3-8B}}&\textbf{60.1}&\textbf{8.0} \\
         \midrule
         Qwen2.5-7B&41.1&30.3\\
         Qwen2.5-7B\textsubscript{+Depth+REC}&49.4&11.8\\
         \textcolor{teal}{\textit{\textbf{Ego3D-VLM}} \textsubscript {Qwen2.5-7B}}&\textbf{54.3}&\textbf{9.5} \\
         \bottomrule
    \end{tabular}}
    \label{tab:tool_experiment}
    \vspace{-8pt}
\end{wraptable}
This category enhances the base VLM with a metric depth estimator and a REC model. We pass each query to the REC model \citep{groundingdino2023hf} to extract BBox of the referred objects, and use a metric depth estimator \citep{depthanythingv2hf} to estimate depth values. We then construct a list that pairs each matched referring expression with its corresponding BBox and estimated depth (i.e., distance from the ego). An example entry in this list is:
{\texttt{[Front-View: Detected pedestrian with red hat at bbox [x1, y1, x2, y2], depth: {z}, Back-View:...]}} (examples in the Appendix \ref{ssec:app_prompts}). 
As shown in Table~\ref{tab:tool_experiment}, equipping the VLM with REC and depth estimator models improves its baseline performance, highlighting the base model's limitations in accurately identifying objects and estimating their depths. However, even with these enhancements, it falls short compared to our proposed \textit{Ego3D-VLM} framework. This underscores the importance of integrating depth information into a unified map representation, enabling the VLM to reason more effectively about the 3D relationships between objects across different views.

\subsection{Ego3D-VLM on Further Benchmarks}
\vspace{-5pt}
Our \textit{Ego3D-VLM} is designed for ego-centric multi-view cases, which are particularly relevant for AI agent applications. In this section, however, we evaluate the performance of \textit{Ego3D-VLM} in alternative multi-view settings, specifically those used in All-Angle Bench \citep{all-angle} and VSI-Bench \citep{vis-bench}. Table \ref{tab:all-angle} presents the performance of \textit{Ego3D-VLM} {compared to open- and closed-source baseline VLMs} on All-Angle Bench and VSI-Bench. Despite the differences in data settings compared to our primary benchmark, \textit{Ego3D-VLM} still outperforms the respective baselines, demonstrating its adaptability across diverse multi-view scenarios. More details are given in the Appendix \ref{ssec:app_inference_time}.
\setlength{\tabcolsep}{3pt} 
\setlength{\intextsep}{3pt}
\renewcommand{\arraystretch}{0.5}
\begin{wraptable}{r}{0.48\textwidth}
  \vspace{+3pt}
  \centering
  \caption{Comparison results on All-Angle Bench and VSI-Bench (accuracy).}
  \footnotesize
  \resizebox{0.48\textwidth}{!}{
        \begin{tabular}{l|cc}
        \toprule \rowcolor[gray]{0.9}
             Model & All-Angle-Bench & VSI-Bench  \\ \midrule
             GPT-4o & 47.8 & 34.0\\
             Gemini-1.5-Pro & 47.4 & 45.4 \\
             Gemini-1.5-Flash & 46.6 & 42.1 \\ \midrule
             InternVL3-8B & 47.9 & 38.1\\ 
             \textcolor{teal}{\textit{\textbf{Ego3D-VLM}}\textsubscript {InternVL3-8B}} & \textbf{49.5} &\textbf{39.6}\\ \cmidrule(lr){2-3}
             InternVL3-14B & 50.3 & 38.2 \\ 
             \textcolor{teal}{\textit{\textbf{Ego3D-VLM}}\textsubscript {InternVL3-14B}} & \textbf{52.1}  & \textbf{40.0} \\ %\cmidrule(lr){2-3}
             %InternVL3-38B & 54.4 & 41.1 \\  
             %\textcolor{teal}{\textit{\textbf{Ego3D-VLM}}\textsubscript {InternVL3-38B}} &\textbf{55.7} & \textbf{42.4} \\
             \bottomrule
        \end{tabular}}    
    \label{tab:all-angle}
%\vspace{1pt}
\end{wraptable}
% \vspace{-5pt}
\setlength{\tabcolsep}{3pt} 
\renewcommand{\arraystretch}{0.5}
\begin{wraptable}{l}{0.48\textwidth}
  \vspace{-30pt}
  \centering
  \caption{Ablation on \textit{Ego3D-VLM} main components. v\textsubscript{4}: all components.}
    %v\textsubscript{5} and v\textsubscript{6} are shown for further analysis.}
  \footnotesize
  \resizebox{0.48\textwidth}{!}{
    \begin{tabular}{p{0.15cm}l|cc}
    \toprule
        \rowcolor[gray]{0.94}
         &&{{Mult. Choice}} &{{Abs. Ans.}}\\ \rowcolor[gray]{0.94}
         &&(Acc. $\uparrow$)&(RMSE$\downarrow$)\\ \midrule
         v\textsubscript{0}&InternVL3-8B &{43.1}&{27.2} \\ 
         v\textsubscript{1}&\hspace{1mm} + CogMap (est. $R,T, K$) &{56.0}&{10.8}\\
         v\textsubscript{2}&\hspace{1mm} +  $K$ &{56.3}&{10.1}\\
         v\textsubscript{3}&\hspace{1mm} + $R,T$   &{58.4}&{10.4}\\ \rowcolor[gray]{0.95}
         v\textsubscript{4}&\hspace{1mm} + Scaling &{{60.1}}&{{8.0}} \\ \hline
         v\textsubscript{5}&\hspace{1mm} + List of objects  & {61.8}&{6.5}\\
         % v\textsubscript{5}&\hspace{1mm} + GT XY & \\
         v\textsubscript{6}&\hspace{1mm} GT CogMap &{79.4}&{1.3}\\
         \bottomrule
    \end{tabular}}
    \label{tab:ablation}
%\vspace{-3pt}
\end{wraptable}
\setlength{\tabcolsep}{3pt} 
\renewcommand{\arraystretch}{0.5}
\begin{wraptable}{r}{0.48\textwidth}
    \vspace{-5pt}
    \centering
    \caption{Perception-reasoning disentanglement.}
    \footnotesize
    \resizebox{0.48\textwidth}{!}{
    \begin{tabular}{l|c|cc}
        \toprule \rowcolor[gray]{0.94}
         &GT&{{Mult. Choice}} &{{Abs. Ans.}}\\ \rowcolor[gray]{0.94}
         Model&BBOX&(Acc. $\uparrow$)&(RMSE$\downarrow$)\\ \midrule
         InternVL3-8B&&43.1&27.2\\
         InternVL3-8B &\checkmark &\textbf{50.2}&\textbf{21.0}\\ \midrule
         Ego3D-VLM\textsubscript{InternVL3-8B} &&60.1&8.0\\
         Ego3D-VLM\textsubscript{InternVL3-8B}&\checkmark &\textbf{62.2}&\textbf{6.8}\\ \midrule
         InternVL3-14B&&51.7&17.6\\
         InternVL3-14B &\checkmark &\textbf{51.8}&\textbf{16.2}\\ \midrule
         Ego3D-VLM\textsubscript{InternVL3-14B} &&66.5&7.7\\
         Ego3D-VLM\textsubscript{InternVL3-14B} &\checkmark &\textbf{66.8}&\textbf{7.5}\\
         \bottomrule
    \end{tabular}}
    \label{tab:disantanglment}
    %\vspace{-3pt}
\end{wraptable}
% \vspace{-25pt}
\subsection{Ablation Studies}
\vspace{-5pt}
\textbf{\textit{Ego3D-VLM} Components.} Table \ref{tab:ablation} presents an ablation on \textit{Ego3D-VLM} core components. Starting from the baseline, v\textsubscript{0}, we incrementally add each component. {In v\textsubscript{1}, we use a cognitive map with estimated rotation ($R$), translation ($T$), and intrinsic params ($K$). 
Specifically, all cameras are positioned at the coordinate center with $T = [0, 0, 0]$. The front camera uses an identity rotation matrix, while all the other cameras---front-right, right, back-right, back, etc.
---are incrementally rotated {45\textdegree}~around the Y-axis. The focal length is estimated from the images’ approximate field of view.}
In v\textsubscript{2}, we use the actual $K$ and in v\textsubscript{3} we further use actual $R$, and $T$ relative to the front camera, i.e., often available as a fixed parameter for an embodied AI agent. v\textsubscript{3} and v\textsubscript{1} obtain a comparable RMSE, while v\textsubscript{3} is only 2.4\% higher in multi-choice QAs. 
Thus, even with estimated camera parameters our cognitive map can significantly boost the baseline's performance. v\textsubscript{4} adds relational scaling to cognitive map which decreases the RMSE by 2.5 meters. 
v\textsubscript{4} is indeed \textit{Ego3D-VLM} with all components. We provide two more ablations to evaluate the upper-bound of \textit{Ego3D-VLM}. v\textsubscript{5} assumes that the list of objects (only their names) in the input query are provided which enhances the REC results. 
v\textsubscript{6} is the ground-truth cognitive map with ground-truth 3D locations.The difference between v\textsubscript{6} and human level is only 5\% showing the upper bound of \textit{Ego3D-VLM}.

\setlength{\tabcolsep}{3pt} 
\renewcommand{\arraystretch}{0.5}
\begin{wraptable}{r}{0.43\textwidth}
    \vspace{-4pt}
    \centering
    \footnotesize
    \caption{Ablation on cognitive map format.}
    \resizebox{0.40\textwidth}{!}{
    \begin{tabular}{l|cc}
        \toprule \rowcolor[gray]{0.94}
         &{{Mult. Choice}} &{{Abs. Ans.}}\\ \rowcolor[gray]{0.94}
         &(Acc. $\uparrow$)&(RMSE$\downarrow$)\\ \midrule
         Visual Cog-Map&50.9&14.4\\
         JSON Cog-Map&60.0&8.4\\
         Textual Cog-Map&\textbf{60.1}&\textbf{8.0} \\
         \bottomrule
    \end{tabular}}
    \label{tab:cogmap_format}
    %\vspace{-10pt}
\end{wraptable}
\textbf{Perception-Reasoning Disentanglement.}
{In order to disentangle perception from reasoning, we add the ground-truth 2D BBox for all objects to the benchmark. 
This allows repeating our experiments with BBox overlaid, effectively isolating the reasoning and perception.
As expected, adding GT BBox as visual prompts to the images will help the VLM to handle perception easier, therefore improving the results.
The results in Table \ref{tab:disantanglment} show that the major limitation of baselines is not 2D perception, but 3D perception and reasoning. More reasoning analysis on \texttt{Ego3D-Bench} as well as \textit{Ego3D-VLM}'s robustness in challenging conditions are given in Appendix \ref{ssec:app_thinking} and \ref{ssec:app_robustness}.}

\textbf{Cognitive Map Format.} We explore 3 different formats for our generated cognitive map: visual, JSON, and textual. Examples of the visual and textual maps are shown in Figure \ref{fig:cog_map}. The average results are presented in Table \ref{tab:cogmap_format}. Among the three, the textual and JSON formats achieve the best overall performance.

\vspace{-5pt}
\section{Conclusion and Future Work}
\vspace{-5pt}
In this work, we proposed \texttt{Ego3D-Bench}, a benchmark for spatial understanding of VLMs on ego-centric multi-view images. Overall, the benchmark shows a significant gap between human scores and VLMs. To address this limitation, we provided a post-training solution named \textit{Ego3D-VLM} to enhance the performance of VLMs. Future work should explore fine-tuning VLMs with ego-centric multi-view QAs and incorporating 3D projection modules proposed in \textit{Ego3D-VLM} in the course of fine-tuning. The limitations of this work are provided in the Appendix (\ref{ssec:app_limitation}).

\medskip
%\clearpage
{
    \small
    \bibliographystyle{ieeenat_fullname}
    \bibliography{main}
}

\clearpage
\appendix
\section{Appendix}
In this appendix, we present the following additional discussions and experimental results corresponding to our proposed \texttt{Ego3D-Bench} and \textit{Ego3D-VLM}:
\begin{itemize}
    %\item Code and \texttt{Ego3D-Bench} 
    \item \texttt{Ego3D-Bench} vs. Other Benchmarks
    \item Further details on \texttt{Ego3D-Bench} creation and question templates
    \item Benchmarking the thinking of \texttt{Ego3D-Bench} as an Open-QA
    \item Results of the three source datasets: NuScenes, Waymo, and Argoverse     
    \item Benchmarking more generalist VLMs    
    \item Inference time analysis of \textit{Ego3D-VLM} vs. other baselines
    \item Robustness of \textit{Ego3D-VLM} in challenging conditions
    \item Qualitative results and prompts structure
    \item Limitations and future work
\end{itemize}

% \subsection{Code}
% In order for our results to be reproducible, we share our code as supplementary materials, with detailed instructions included in the associated README.md file. We have also included the \texttt{Ego3D-Bench} in the supplementary material. However, due to size limitation, images should be downloaded from the source datasets. Instructions to download images and models are provided in the README.md file. 

\subsection{\texttt{Ego3D-Bench} vs. Other Benchmarks}
\label{ssec:app_prompts}
\begin{figure}[h!]
    \centering            
    \includegraphics[width=\linewidth]{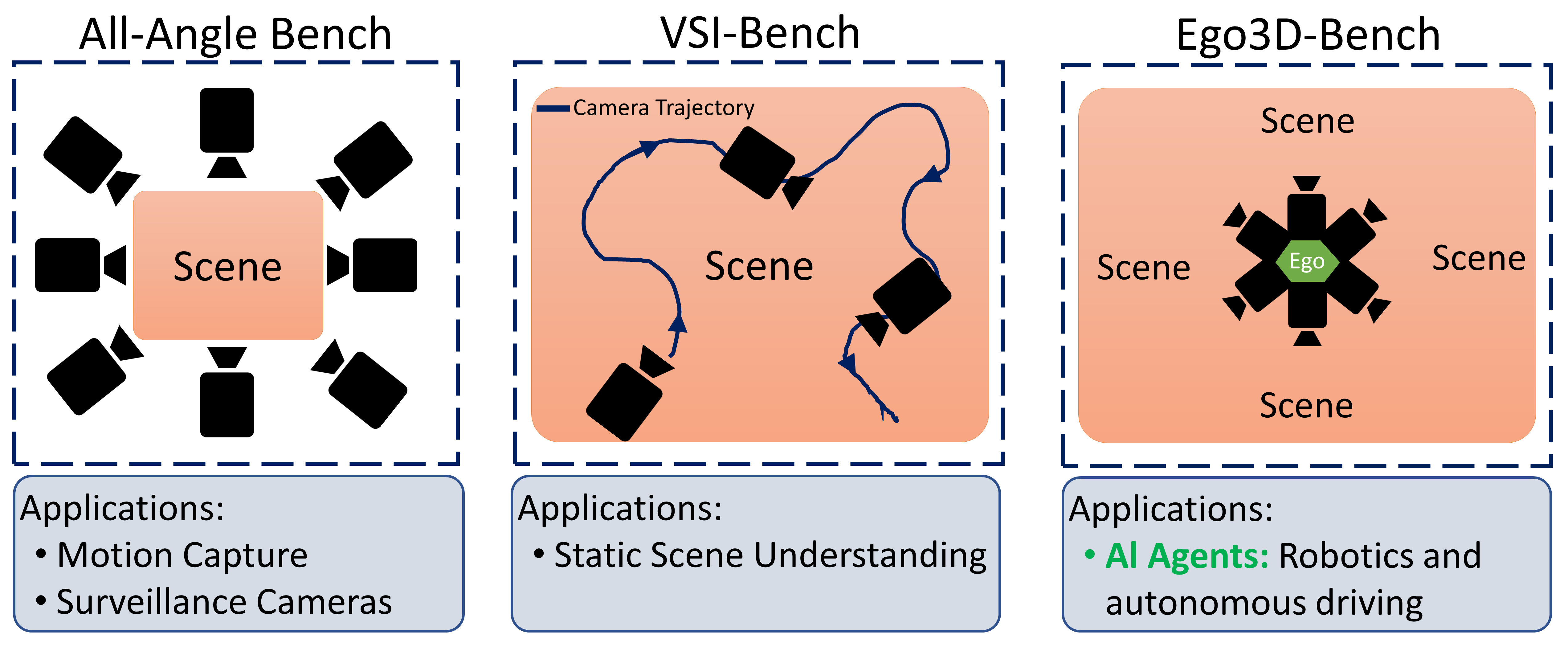}
    \caption{Settings of multi-view/video spatial understanding  benchmarks for VLMs.}
    \label{fig:all-angle}
\end{figure}

As described in the main body of this paper, our proposed \textit{Ego3D-VLM} is designed for ego-centric multi-view scenarios, which are particularly relevant for AI agent applications. \textit{Ego3D-VLM} is different from alternative multi-view settings, specifically those used in All-Angle Bench \citep{all-angle} and VSI-Bench \citep{vis-bench}. Figure \ref{fig:all-angle} illustrates the configurations of these benchmarks in comparison to \texttt{Ego3D-Bench}. All-Angle Bench features stationary multi-view cameras observing the same scene from different angles—a setup commonly used in motion capture and surveillance systems. VSI-Bench, on the other hand, involves a single moving camera within a static indoor environment, typically used for static scene reconstruction and SU. In contrast, our \texttt{Ego3D-Bench} assumes multiple cameras mounted on a moving ego agent, capturing the surrounding environment—a configuration suited for AI agent applications such as robotics and autonomous driving.

\subsection{\texttt{Ego3D-Bench}: Further Details}
\label{ssec:app_templates}
\begin{figure}[t!]
    \centering
    \includegraphics[width=1\linewidth]{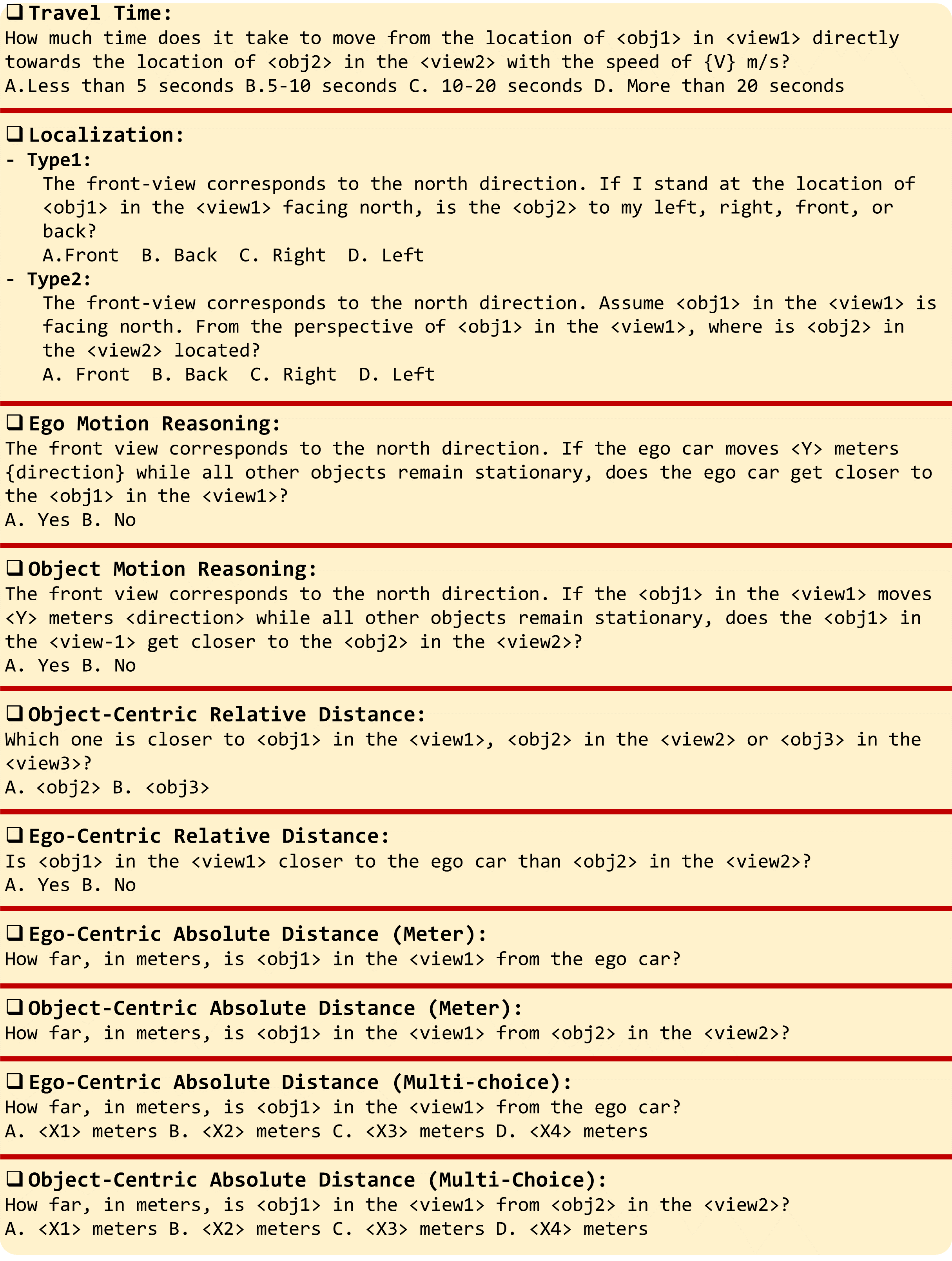}
    \caption{Templates used to create questions of \texttt{Ego3D-Bench}.}
    \label{fig:templates}
\end{figure}

In this section, we provide the templates used to create questions of the benchmark. {Figure}
\ref{fig:templates} shows templates used to create each category of \texttt{Ego3D-Bench}. We replace \texttt{<obj1>}, \texttt{<obj2>}, and \texttt{<obj3>} in the templates with object descriptions. 
\texttt{<view1>}, \texttt{<view2>}, and \texttt{<view3>} are replaced with the camera views from which the object is visible (e.g., Front-Right, Left, etc.).
The placeholder \texttt{<direction>} in motion reasoning tasks is substituted with one of the four cardinal directions—"north", "east", "west", or "south"—and this process is repeated until all directions have been used. The placeholder \texttt{<Y>} is for motion reasoning  tasks and is replaced with a number between 2 to 5 meters. \texttt{<X1>}, \texttt{<X2>}, \texttt{<X3>}, and \texttt{<X4>} serve as placeholders for multiple-choice options in absolute distance estimation tasks. One of these options is replaced with the ground-truth distance, while the remaining are filled with randomly generated values, ensuring that the distance between any two options is at least 8 meters.

\subsection{Benchmarking the Thinking of \texttt{Ego3D-Bench} as an Open-QA}
\label{ssec:app_thinking}
{All our question-answer pairs include a "thinking" step in open-ended format. In this section, we generate "GT thinking" for further open-ended type evaluation. 
To this end, we used the GT cognitive maps along with GPT-4o to generate the reasoning for two categories that require numerical answers: Ego-Centric / Object-Centric Absolute Distance (in meters). Table \ref{tab:open-qa} shows that Ego3D-VLM thinking is better than  the baselines.}

\setlength{\tabcolsep}{3pt} 
\renewcommand{\arraystretch}{0.5}
\begin{table}[h!]
    \centering
    \footnotesize
    \caption{Benchmarking the thinking of VLMs on \texttt{Ego3D-Bench} as an open-ended category. The GT thinking is generated using GT cognitive maps along with GPT-4o.}    
    \begin{tabular}{l|cc}
        \toprule \rowcolor[gray]{0.94}
         &{\textbf{Thinking of Obj. Abs. Dist.}} &{\textbf{Thinking of Ego Abs. Dist.}}\\ \rowcolor[gray]{0.94}
         &GPT-4o Score (0-10)$\uparrow$& GPT-4o Score (0-10)$\uparrow$\\ \midrule
         InternVL3-8B&1.9&2.0\\
         Ego3D-VLM \textsubscript{InternVL3-8B}&\textbf{4.7}&\textbf{5.2}\\ \midrule
         InternVL3-14B &2.3&2.4\\
         Ego3D-VLM \textsubscript{InternVL3-14B}&\textbf{5.7}&\textbf{5.0}\\ \midrule
         InternVL3-38B &2.1&2.5\\ 
         Ego3D-VLM \textsubscript{InternVL3-38B}&\textbf{7.1}&\textbf{3.7}\\ \midrule
         InternVL3-78B &2.1&1.7\\
         Ego3D-VLM \textsubscript{InternVL3-78B}&\textbf{7.0}&\textbf{6.8}\\
         \bottomrule
    \end{tabular}
    \label{tab:open-qa}
\end{table}

\subsection{Benchmarking Generalist VLMs: Further Results}
\label{ssec:app_more_vlms}
In the main paper, we reported results for InternVL3, Qwen2.5-VL, and Ovis-2 as representative open-source models, as they are, to the best of our knowledge, the current SOTA among open-source VLMs. Table \ref{tab:further_quant_results} extends this comparison by including additional generalist VLMs on \texttt{Ego3D-Bench}, such as Gemini-1.5-Flash \citep{google2023gemini}, Gemini-2-Flash \citep{google2023gemini}, Gemini-2.5-Flash \citep{google2023gemini}, Phi-3.5 \citep{phi3}, LLaVA-One-Vision-7B \citep{li2024llava}, LLaVA-Next-Video-7B \citep{zhang2024llavanextvideo}, and DeepSeek-VL2 \citep{deepseek-vl2}. We observe that LLaVA-One-Vision, Phi-3.5, and DeepSeek-VL2 underperform compared to InternVL3 and Qwen2.5-VL models of similar size. For instance, InternVL3-8B achieves an accuracy of 43\%, while LLaVA-One-Vision-7B achieves 38.7\%. Likewise, Phi-3.5 (3.8B) attains an accuracy of 40.3\%, compared to 41\% from Ovis-2 (4B).

\setlength{\tabcolsep}{3pt} 
\begin{table*}[h!]
    \centering
    \footnotesize
    %\vspace{-3pt}
    \caption{     
    {Results of further generalist VLMs on \texttt{Ego3D-Bench}.}
    }   
    \begin{tabular}{l|cccccccc|p{0.65cm}|cc|p{0.65cm}}
          \toprule
          & \multicolumn{9}{c|}{Accuracy (\%) $\uparrow$}&\multicolumn{3}{c}{RMSE$ \downarrow$} \\ \hline \rowcolor[gray]{0.94}
        \rule{0pt}{10pt}& 
         \rotatebox{0}{Ego} &
         \rotatebox{0}{Obj.} &
         \rotatebox{0}{} &
         \rotatebox{0}{Ego} &
         \rotatebox{0}{Obj.} &
         \rotatebox{0}{Travel} &
         \rotatebox{0}{Ego} &
         \rotatebox{0}{Obj.} &
         \rotatebox{0}{} &
         \rotatebox{0}{Ego} &
         \rotatebox{0}{Obj.} &
         \rotatebox{0}{}\\ \rowcolor[gray]{0.94}
         Model&\rotatebox{0}{Dist.} &
         \rotatebox{0}{Dist.} &
         \rotatebox{0}{Loc.} &
         \rotatebox{0}{Mot.} &
         \rotatebox{0}{Mot.} &
         \rotatebox{0}{Time} &
         \rotatebox{0}{Rel.} &
         \rotatebox{0}{Rel.} &
         \rotatebox{0}{\textcolor{blue}{\textbf{Avg.$\uparrow$}}} &
         \rotatebox{0}{Dist.} &
         \rotatebox{0}{Dist.} &
         \rotatebox{0}{\textcolor{blue}{\textbf{Avg.$\downarrow$}}}\\ 
          \midrule
          \rowcolor[gray]{0.9}
          \multicolumn{13}{l}{\textit{ Gemini}} \\
          \hspace{2.5mm} Gemini-1.5 Flash & 40.4& 	26.9& 	50.5& 	70.1& 	61.8& 	28.0& 	53.6& 	75.6& 	\textcolor{blue}{50.9}& 	10.8& 	31.0& 	\textcolor{blue}{20.9} \\
          \hspace{2.5mm} Gemini-2-Flash & 43.3 & 38.6 & 54.1 & 58.0 & 39.3 & 29.1 & 47.1 & 68.5 & \textcolor{blue}{47.2} & 10.9 & 29.4 & \textcolor{blue}{20.1} \\
          \hspace{2.5mm} Gemini-2.5-Flash & 41.3 & 25.6 & 65.1 & 93.4 & 84.8 & 20.1  & 63.5 & 68.5 & \textcolor{blue}{57.8} & 11.4 & 19.6 & \textcolor{blue}{15.5}\\
          \midrule
          \rowcolor[gray]{0.9}
          \multicolumn{13}{l}{\textit{Phi}} \\
          % \hspace{2.5mm} Phi4& \\ 
          \hspace{2.5mm} Phi3.5&28.3 & 30.8 & 24.2 &59.6 &55.9 & 21.3 & 45.3 & 56.7 & \textcolor{blue}{40.3} & 22.8 & 49.4 & \textcolor{blue}{36.1}\\
          \midrule
          \rowcolor[gray]{0.9}
          \multicolumn{13}{l}{\textit{LLaVA Family}} \\
          \hspace{2.5mm} LLaVA-Next-Video-7B& 30.2& 	26.6& 	42.2& 	60.8& 	58.0& 	48.3& 	55.1& 	73.5& 	\textcolor{blue}{49.3}& 	19.5& 	22.8& 	\textcolor{blue}{21.2} \\ 
          \hspace{2.5mm} LLaVA-OV-7B& 23.5& 	22.1& 	58.4& 	56.3& 	52.5& 	17.9& 	59.0& 	20.0& 	\textcolor{blue}{38.7}& 	17.8& 	19.3& 	\textcolor{blue}{18.5} \\ 
          \midrule
          \rowcolor[gray]{0.9}
          \multicolumn{13}{l}{\textit{DeepSeek Family}} \\
          \hspace{2.5mm} DeepSeek-VL2-tiny& 34.7& 	29.0& 	19.9& 	60.0& 	59.3& 	46.6& 	57.8& 65.0& \textcolor{blue}{46.5}&	14.0& 	18.7& 	\textcolor{blue}{16.3} \\
          \hspace{2.5mm} DeepSeek-VL2-Small& 18.1& 	19.2& 	28.6& 	50.3& 	55.6& 	37.4& 	50.5& 	59.3& 	\textcolor{blue}{39.9}& 	12.7& 	14.7& 	\textcolor{blue}{13.7} \\ 
          \hspace{2.5mm} DeepSeek-VL2& 22.3& 	25.3& 	32.4& 	60.0& 	59.1& 	21.4& 	56.7& 	62.0& 	\textcolor{blue}{42.4}& 	20.3& 	17.4& 	\textcolor{blue}{18.8} \\
         \bottomrule
    \end{tabular} 
    %\vspace{-7pt}
    \label{tab:further_quant_results}
\end{table*}

\subsection{Inference Time Analysis}
\label{ssec:app_inference_time}
Table \ref{tab:inference_time} shows the inference time analysis of \textit{Ego3D-VLM} compared to baseline models. We report End-to-End Latency (E2E Lat.) in seconds and Peak Memory in GB. For the E2E Lat., we measure the average end-to-end inference of models on 50 samples of \texttt{Ego3D-Bench} and for the peak memory we report the peak memory usage during inference on the same samples. Experiments are performed using flash-attention-2. The memory and latency overhead of \textit{Ego3D-VLM} over InternVL3-78B is 0.6\% and 31\% , respectively. The main reason for the latency overhead is that the model reasons more when cognitive map is provided in the \textit{Ego3D-VLM}. In order to make the inference more efficient and deal with the latency overhead of the VLMs, post-training techniques such as quantization \citep{casp}, token pruning \cite{divprune}, or knowledge distillation \citep{gold} can be used.
\begin{table}[h!]
    \centering
    % \footnotesize
    \caption{Inference time and memory usage of \textit{Ego3D-VLM} compared to the baselines.}
    \begin{tabular}{lcc||lccc}
    \toprule \rowcolor[gray]{0.9}
          &E2E Lat.& Memory& &E2E Lat.& Memory \\ \rowcolor[gray]{0.9}
          &(sec) & (GB)& &(sec) & (GB)\\ \midrule
         InternVL3-8B & 5.2&18.1&InternVL3-38B&16.9&80.0\\
         \textcolor{teal}{{\textit{\textbf{Ego3D-VLM}}\textsubscript {InternVL3-8B}}}&8.6&26.5 &
         \textcolor{teal}{\textit{\textbf{Ego3D-VLM}}\textsubscript {InternVL3-38B}}&19.1&84.6 \\ \midrule
         InternVL3-14B & 15.5&33.1&InternVL3-78B&35.0&161.7\\
         \textcolor{teal}{{\textit{\textbf{Ego3D-VLM}}\textsubscript {InternVL3-14B}}} &16.4&40.2 &\textcolor{teal}{{\textit{\textbf{Ego3D-VLM}}\textsubscript {InternVL3-78B}}}&46.9& 162.4\\
    \bottomrule
    \end{tabular}
    \label{tab:inference_time}
\end{table}

\subsection{Robustness of \textit{Ego3D-VLM} in Challenging Conditions}
\label{ssec:app_robustness}
To isolate tool performance, we included an ablation with ground-truth cognitive maps in the main body of the paper (Table \ref{tab:ablation}), which improve the results from 60.1\% to 79.4\% accuracy. This confirms that, there is still a gap between the accuracy of such imperfect tools and ground-truth REC/depth. However, compared to the baseline, our solution still achieves more robust results even in challenging conditions such as low brightness, motion blur, and occlusion. To support this claim, we simulated these conditions and re-ran more specific experiments shown in the Table \ref{tab:robustness}. 
Although G-DINO and Depth-Anything used in Ego3D-VLM are among the best tools for REC and depth estimation, our solution is orthogonal to any such tools, and has no limitation in this regard.

\setlength{\tabcolsep}{3pt} 
\renewcommand{\arraystretch}{0.5}
\begin{table}[h!]
    \centering
    \footnotesize
    \caption{Robustness of external tools in scenarios involving occlusion, motion blur, and low light.}    
    \begin{tabular}{l|cc}
        \toprule \rowcolor[gray]{0.94}
         &{\textbf{Mult. Choice}} &{\textbf{Abs. Ans.}}\\ \rowcolor[gray]{0.94}
         &(Acc. $\uparrow$)&(RMSE$\downarrow$)\\ \midrule
         InternVL3-8B&43.1&27.2\\
         Ego3D-VLM \textsubscript{InternVL3-8B}&\textbf{60.1}&\textbf{8.0}\\ \midrule
         InternVL3-8B [60\% Darkness]&41.1&29.8 \\
         Ego3D-VLM \textsubscript{InternVL3-8B} [60\% Darkness]&\textbf{59.6}&\textbf{10.6}\\ \midrule
         InternVL3-8B [Motion Blur, 15×1 kernel]&42.5&28.5 \\ 
         Ego3D-VLM \textsubscript{InternVL3-8B} [motion blur, 15×1 kernel]&\textbf{57.9}&\textbf{9.9}\\ \midrule
         InternVL3-8B [30\% Occlusion]&42.0&28.9 \\
         Ego3D-VLM \textsubscript{InternVL3-8B} [30\% Occlusion]&\textbf{58.7}&\textbf{10.7}\\
         \bottomrule
    \end{tabular}
    \label{tab:robustness}
\end{table} 

\subsection{Results of different Source Datasets}
Table \ref{tab:source_split} presents the results for different source dataset splits used to create \texttt{Ego3D-Bench}. Despite the varying number of camera viewpoints across the three datasets, the performance deviations are minimal. This consistency underscores the reliability of \texttt{Ego3D-Bench} as a benchmark, indicating that model performance is not heavily influenced by the specific choice of source data—a desirable property for robust and fair benchmark design.

\begin{table*}[h!]
    \centering
    \footnotesize
    %\vspace{-3pt}
    \caption{     
    {Results on samples of \texttt{Ego3D-Bench} generated from each of the source datasets.}
    }   
    \begin{tabular}{l|cc|cc|cc}
          \toprule
          & \multicolumn{2}{c|}{NuScenes} & \multicolumn{2}{c|}{Waymo} & \multicolumn{2}{c}{Argoverse} \\ 
          & Acc$\uparrow$ & RMSE$\downarrow$ & Acc$\uparrow$ & RMSE$\downarrow$ & Acc$\uparrow$ & RMSE$\downarrow$\\ 
         \midrule
         \rowcolor[gray]{.90}
         \multicolumn{7}{l}{\textit{Closed-source Models}} \\ % No vertical borders!
         \hspace{2.5mm} GPT-4o & 60.9 &21.2 &60.0 & 19.0 & 59.4 &17.1 \\
          \hspace{2.5mm} \textcolor{teal}{\textit{\textbf{Ego3D-VLM}} \textsubscript{GPT-4o}}&\textbf{73.5}&\textbf{8.5}&\textbf{73.1}&\textbf{7.5}&\textbf{72.7}&\textbf{6.2}\\ \cmidrule(lr){2-7}
         \hspace{2.5mm} Gemini-1.5-Pro &59.3&26.4&55.6&12.7&57.5&19.5 \\
         \hspace{2.5mm} \textcolor{teal}{\textit{\textbf{Ego3D-VLM}} \textsubscript {Gemini-1.5-Pro}} &\textbf{77.1}&\textbf{8.1}&\textbf{68.9}&\textbf{6.3}&\textbf{73.0}&\textbf{7.1}\\
         \midrule
         \rowcolor[gray]{0.9}
         \multicolumn{7}{l}{\textit{Qwen2.5 Family}} \\
         \hspace{2.5mm} Qwen2.5-3B&39.7&33.4&38.6&31.5&38.9&31.2 \\ 
         \hspace{2.5mm} \textcolor{teal}{\textit{\textbf{Ego3D-VLM}} \textsubscript {Qwen2.5-3B}} & \textbf{44.7} & \textbf{13.0} &\textbf{40.3}&\textbf{12.1}&\textbf{48.1}&\textbf{12.6}\\  \cmidrule(lr){2-7}
         \hspace{2.5mm} Qwen2.5-7B&40.9&36.0&40.6&24.7&41.6&30.1\\ 
         \hspace{2.5mm} \textcolor{teal}{\textit{\textbf{Ego3D-VLM}} \textsubscript {Qwen2.5-7B}}&\textbf{56.4}&\textbf{11.3}&\textbf{51.9}&\textbf{8.3}&\textbf{54.4}&\textbf{8.7} \\ \cmidrule(lr){2-7}
         \hspace{2.5mm} Qwen2.5-32B&56.6&17.5&56.0&11.0&59.2&18.7 \\
         \hspace{2.5mm} \textcolor{teal}{\textit{\textbf{Ego3D-VLM}} \textsubscript {Qwen2.5-32B}}&\textbf{64.3}&\textbf{17.1}&6\textbf{4.8}&\textbf{10.6}&\textbf{64.0}&\textbf{12.1} \\ \cmidrule(lr){2-7}
         \hspace{2.5mm} Qwen2.5-72B&58.2&21.2&57.7&11.1&58.2&16.2 \\
         \hspace{2.5mm} \textcolor{teal}{\textit{\textbf{Ego3D-VLM}} \textsubscript {Qwen2.5-72B}}&\textbf{74.4}&\textbf{8.5}&\textbf{64.4}&\textbf{6.4}&\textbf{69.5}&\textbf{7.6}\\ 
         \midrule
         \rowcolor[gray]{0.9}
         \multicolumn{7}{l}{\textit{InternVL3 Family}} \\
         \hspace{2.5mm} InternVL3-8B&42.3&28.3&44.0&21.7&42.9&31.5 \\
         \hspace{2.5mm} \textcolor{teal}{\textit{\textbf{Ego3D-VLM}} \textsubscript {InternVL3-8B}} &\textbf{62.0}&\textbf{10.5}&\textbf{57.1}&\textbf{6.5}&\textbf{61.3}&\textbf{6.7}
\\ \cmidrule(lr){2-7}
         \hspace{2.5mm} InternVL3-14B&51.7&20.4&50.2&15.8&53.3&16.6\\
         \hspace{2.5mm} \textcolor{teal}{\textit{\textbf{Ego3D-VLM}} \textsubscript {InternVL3-14B}}&\textbf{69.0}&\textbf{8.8}&\textbf{63.7}&\textbf{7.5}&\textbf{67.0}&\textbf{6.3}\\ \cmidrule(lr){2-7}
         \hspace{2.5mm} InternVL3-38B&51.2&31.4&50.5&19.4&53.4&25.4\\
         \hspace{2.5mm} \textcolor{teal}{\textit{\textbf{Ego3D-VLM}} \textsubscript {InternVL3-38B}}&\textbf{71.7}&\textbf{8.7}&\textbf{64.3}&\textbf{8.2}&\textbf{67.0}&\textbf{6.3} \\ \cmidrule(lr){2-7}
         \hspace{2.5mm} InternVL3-78B&60.0&16.7&59.7&10.7&60.1&13.7\\ 
         \hspace{2.5mm} \textcolor{teal}{\textit{\textbf{Ego3D-VLM}} \textsubscript{InternVL3-78B}}&\textbf{76.5}&\textbf{8.3}&\textbf{67.0}&\textbf{6.5}&\textbf{71.0}&\textbf{7.5}\\
         \midrule
         \rowcolor[gray]{0.9}
         \multicolumn{7}{l}{\textit{Ovis2 Family}} \\
         \hspace{2.5mm} Ovis2-4B&41.2&29.2&43.0&18.1&40.9&22.5\\
         \hspace{2.5mm} \textcolor{teal}{\textit{\textbf{Ego3D-VLM}} \textsubscript {Ovis2-4B}} &\textbf{47.3}&\textbf{9.8}&\textbf{47.5}&\textbf{7.1}& \textbf{49.7}&\textbf{8.3} \\ \cmidrule(lr){2-7}
         \hspace{2.5mm} Ovis2-8B&41.5&21.4&41.9&20.7&41.0&20.2\\
         \hspace{2.5mm} \textcolor{teal}{\textit{\textbf{Ego3D-VLM}} \textsubscript {Ovis2-8B}}&\textbf{55.6}&\textbf{9.4}&\textbf{55.1}&\textbf{6.8}&\textbf{57.9}&\textbf{7.0}\\ 
         \cmidrule(lr){2-7}
         \hspace{2.5mm} Ovis2-16B&48.6&16.2&45.5&11.9&46.8&12.9\\
         \hspace{2.5mm} \textcolor{teal}{\textit{\textbf{Ego3D-VLM}} \textsubscript {Ovis2-16B}}&\textbf{62.6}&\textbf{8.5}&\textbf{57.9}&\textbf{7.15}&\textbf{62.2}&\textbf{6.5} \\
         \bottomrule
    \end{tabular} 
    %\vspace{-7pt}
    \label{tab:source_split}
\end{table*}

\subsection{Qualitative Results and Prompts Structure}
\label{ssec:app_prompts}
\begin{figure*}
    \centering    
    \includegraphics[width=1\linewidth]{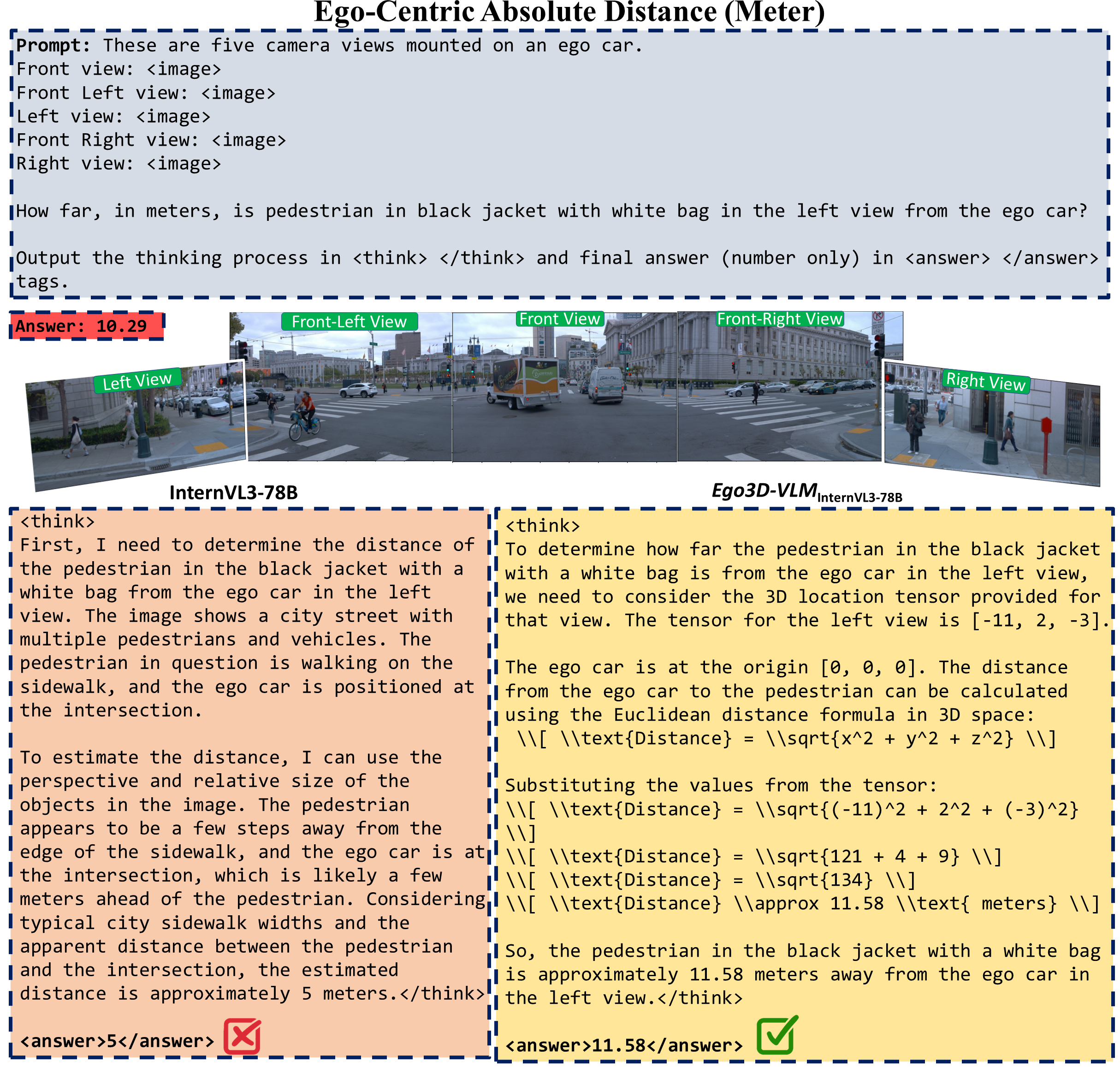}
    \caption{Example responses of the baseline and \textit{Ego3D-VLM} on ego-centric absolute distance task.}
    \label{fig:sample1}
\end{figure*}

\begin{figure*}
    \centering    
    \includegraphics[width=1\linewidth]{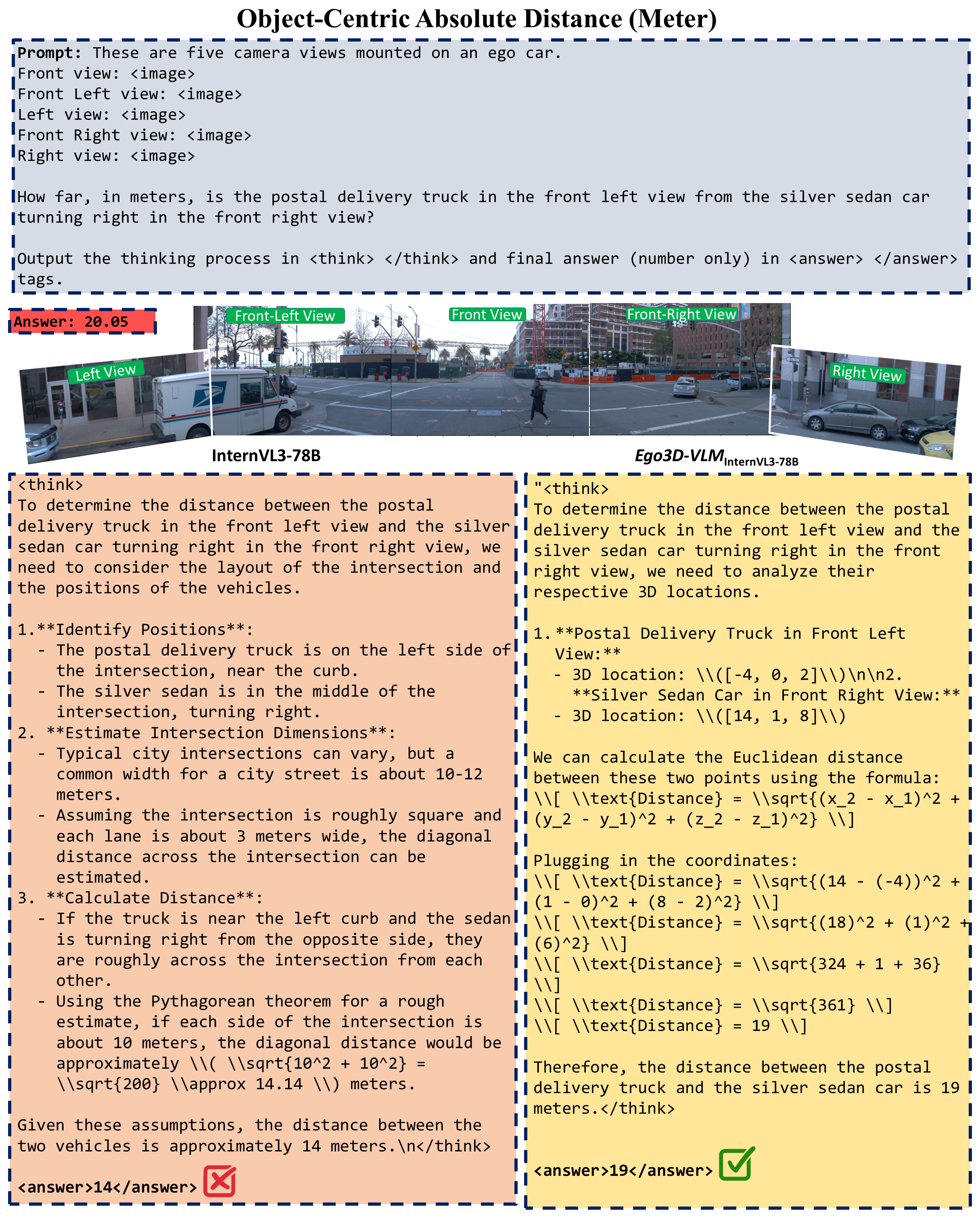}
    \caption{Example responses of the baseline and \textit{Ego3D-VLM} on object-centric absolute distance task.}
    \label{fig:sample2}
\end{figure*}

\begin{figure*}
    \centering    
    \includegraphics[width=1\linewidth]{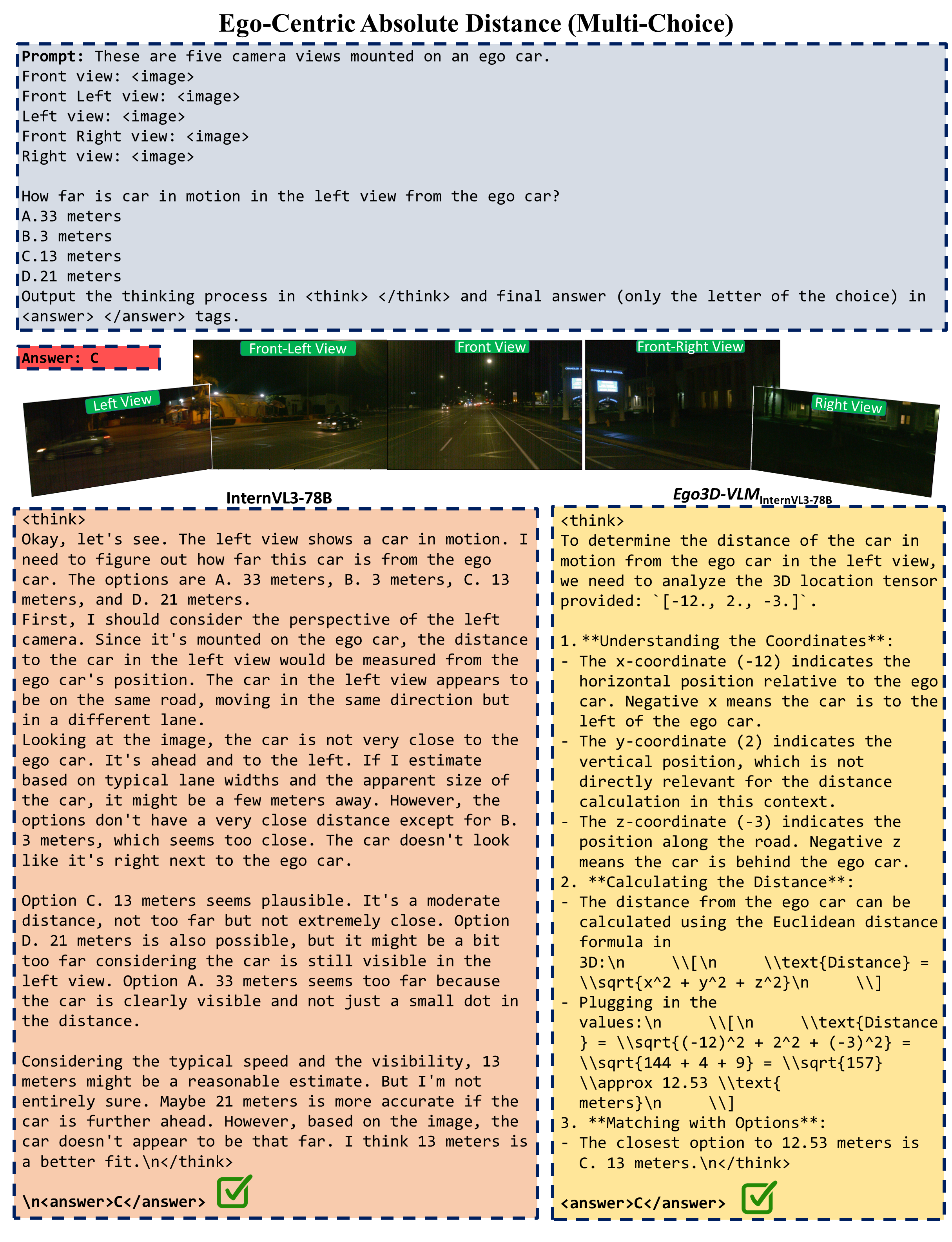}
    \caption{Example responses of the baseline and \textit{Ego3D-VLM} on multi-choice ego-centric absolute  distance task.}
    \label{fig:sample3}
\end{figure*}

\begin{figure*}
    \centering    
    \includegraphics[width=1\linewidth]{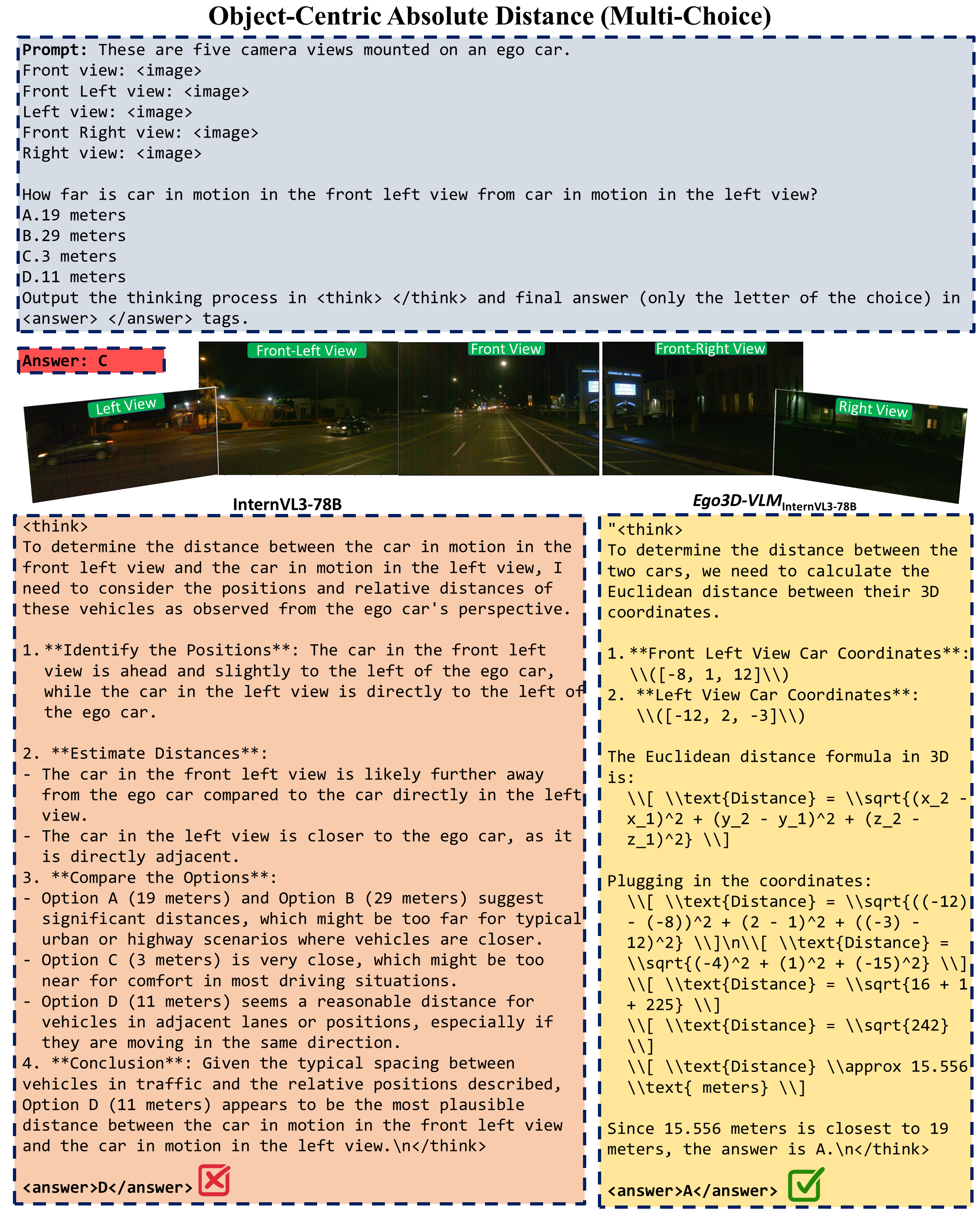}
    \caption{Example responses of the baseline and \textit{Ego3D-VLM} on multi-choice object-centric absolute  distance task.}
    \label{fig:sample4}
\end{figure*}

\begin{figure*}
    \centering    
    \includegraphics[width=1\linewidth]{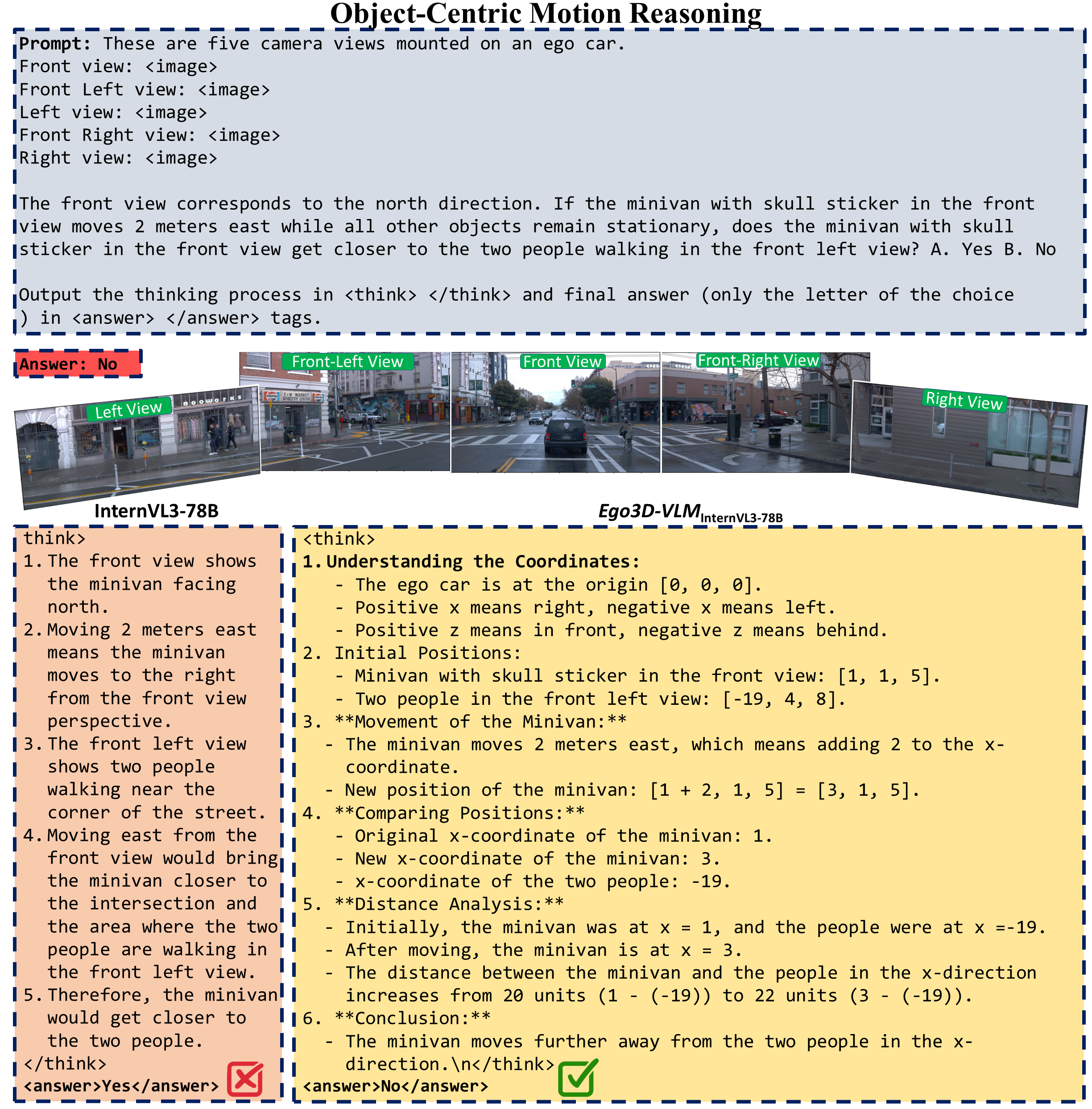}
    \caption{Example responses of the baseline and \textit{Ego3D-VLM} on object-centric motion reasoning task.}
    \label{fig:sample5}
\end{figure*}

\begin{figure*}
    \centering    
    \includegraphics[width=1\linewidth]{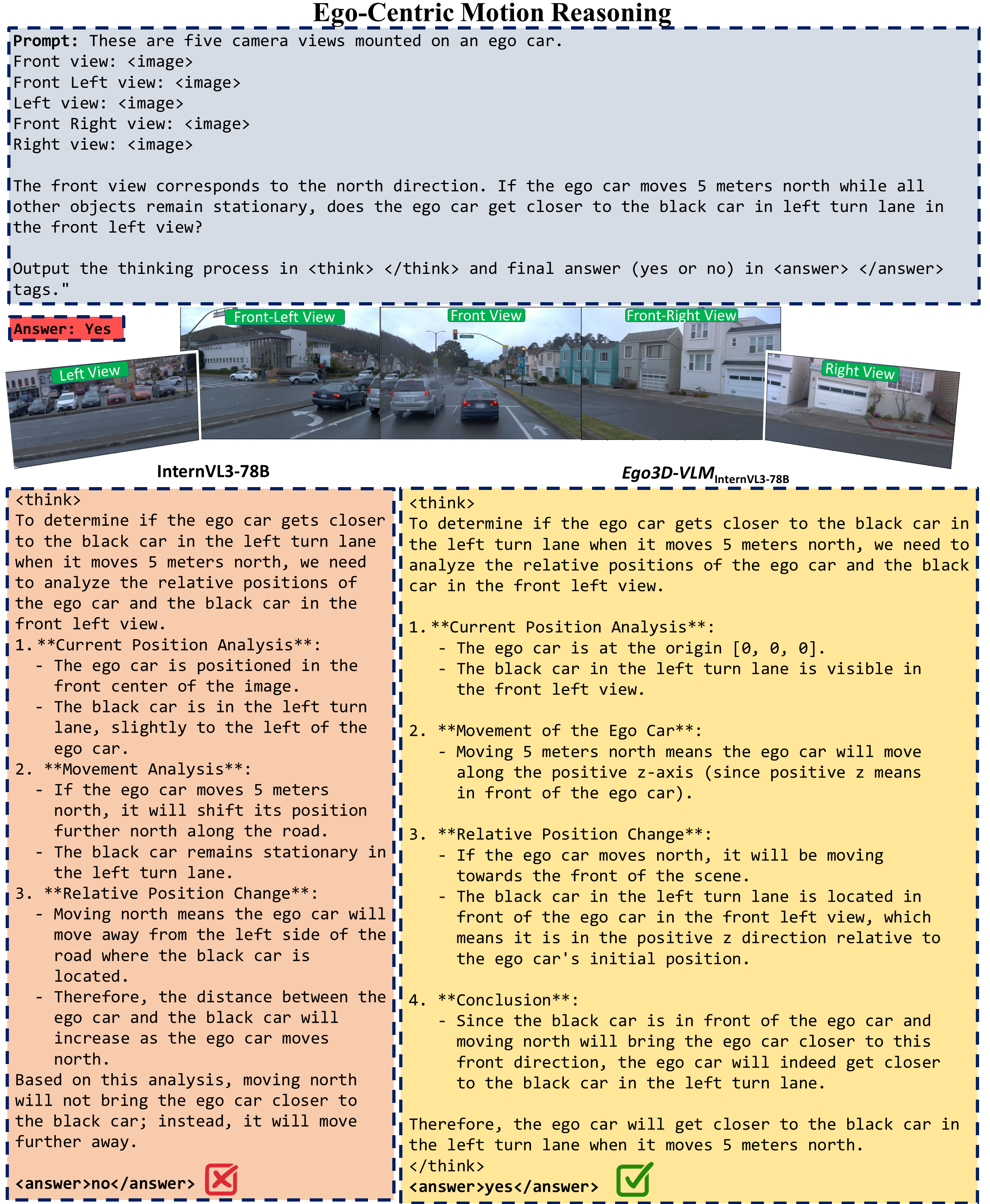}
    \caption{Example responses of the baseline and \textit{Ego3D-VLM} on ego-centric motion reasoning task.}
    \label{fig:sample6}
\end{figure*}

\begin{figure*}
    \centering    
    \includegraphics[width=1\linewidth]{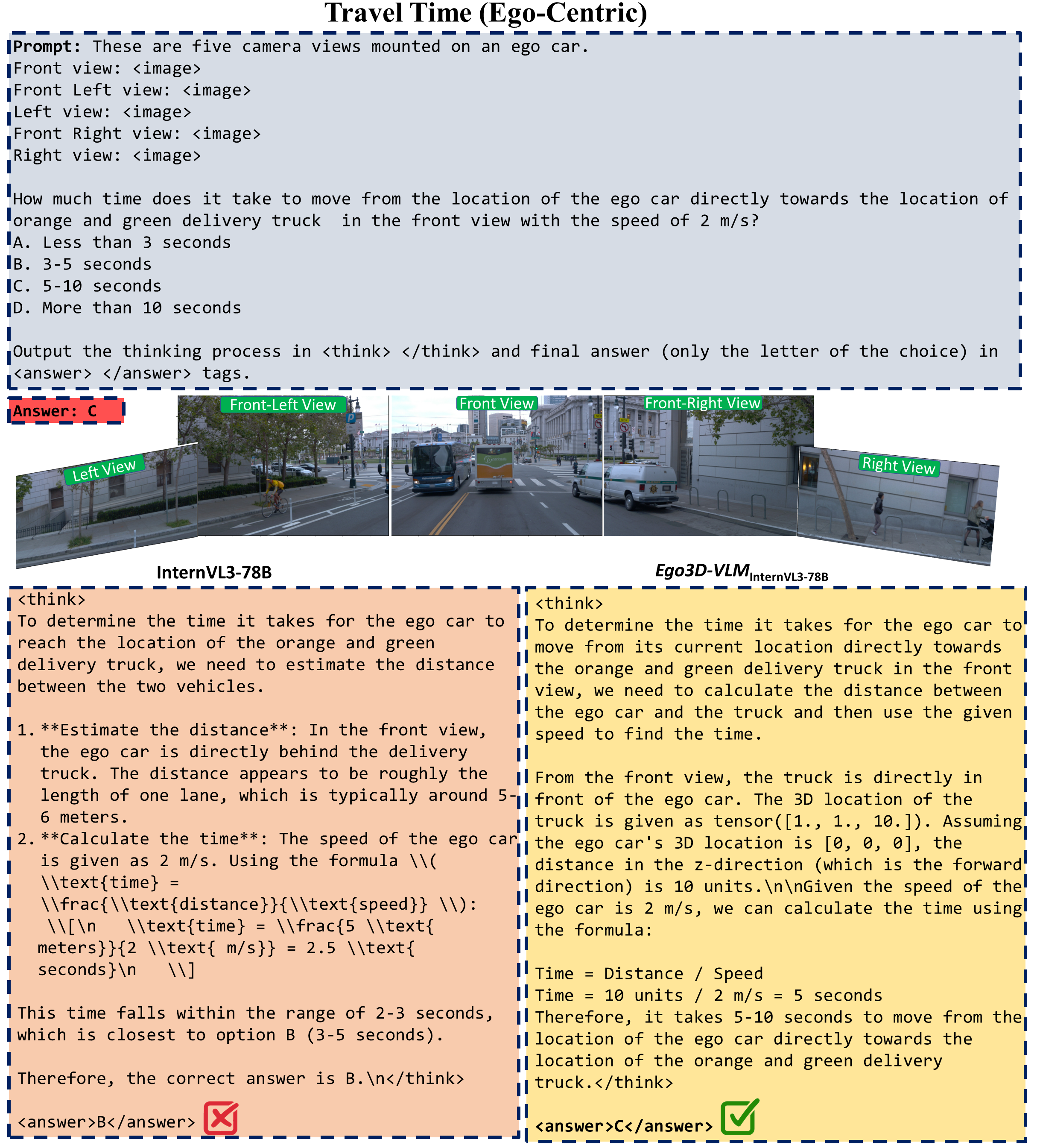}
    \caption{Example responses of the baseline and \textit{Ego3D-VLM} on travel time (ego-centric) task.}
    \label{fig:sample7}
\end{figure*}

\begin{figure*}
    \centering    
    \includegraphics[width=1\linewidth]{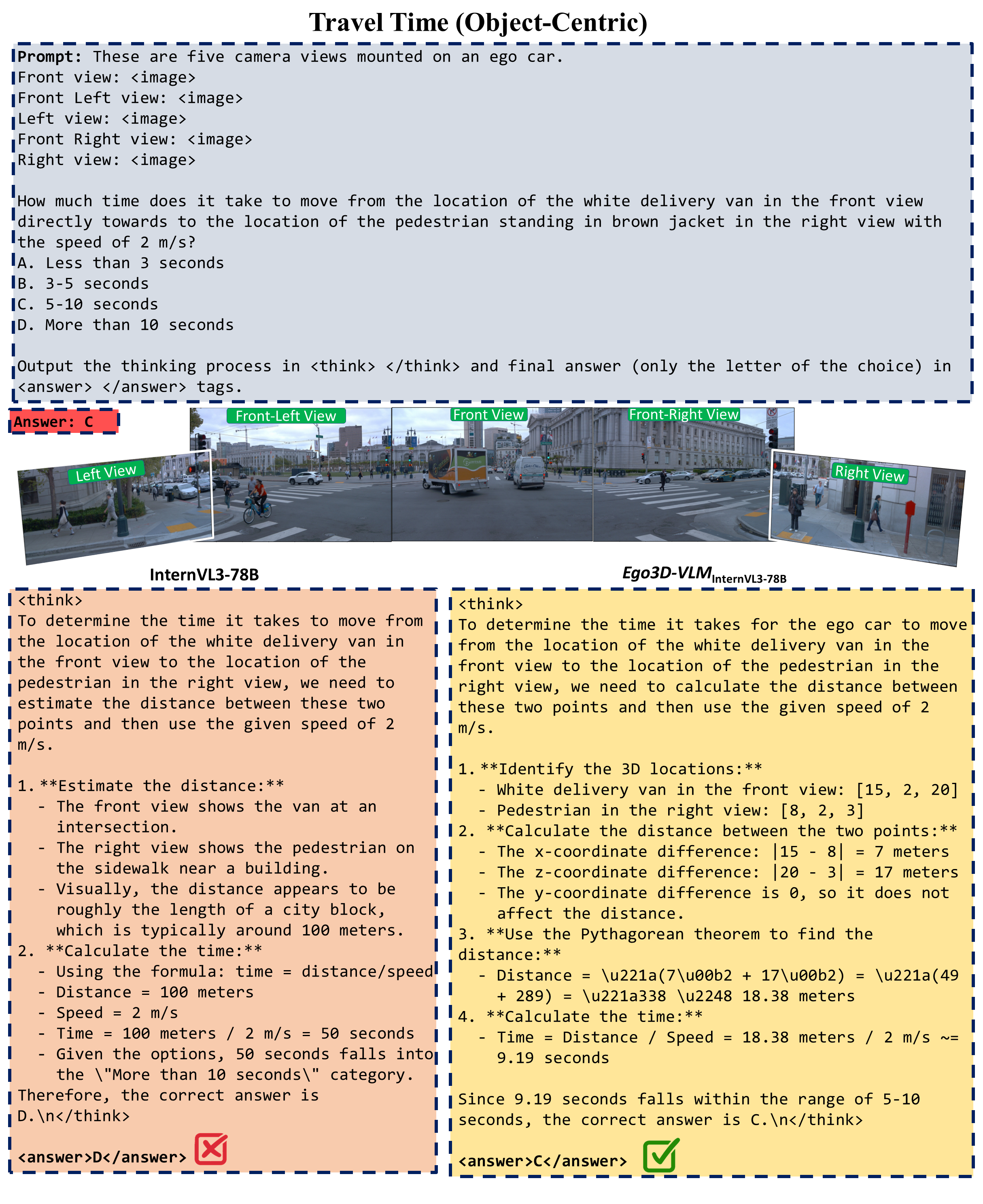}
    \caption{Example responses of the baseline and \textit{Ego3D-VLM} on travel time (object-centric) task.}
    \label{fig:sample8}
\end{figure*}

\begin{figure*}
    \centering    
    \includegraphics[width=1\linewidth]{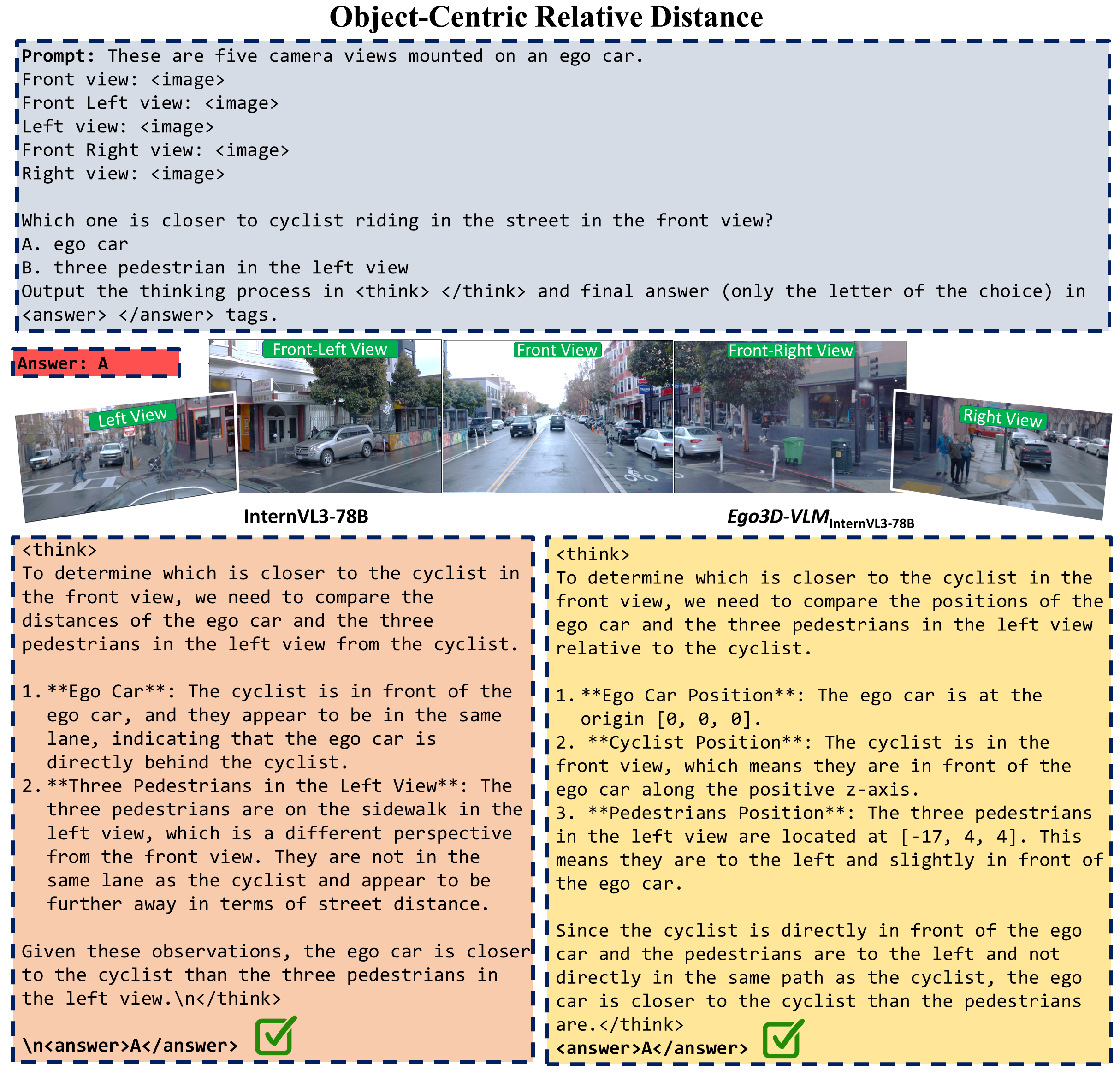}
    \caption{Example responses of the baseline and \textit{Ego3D-VLM} on object-centric relative distance task.}
    \label{fig:sample9}
\end{figure*}

\begin{figure*}
    \centering    
    \includegraphics[width=1\linewidth]{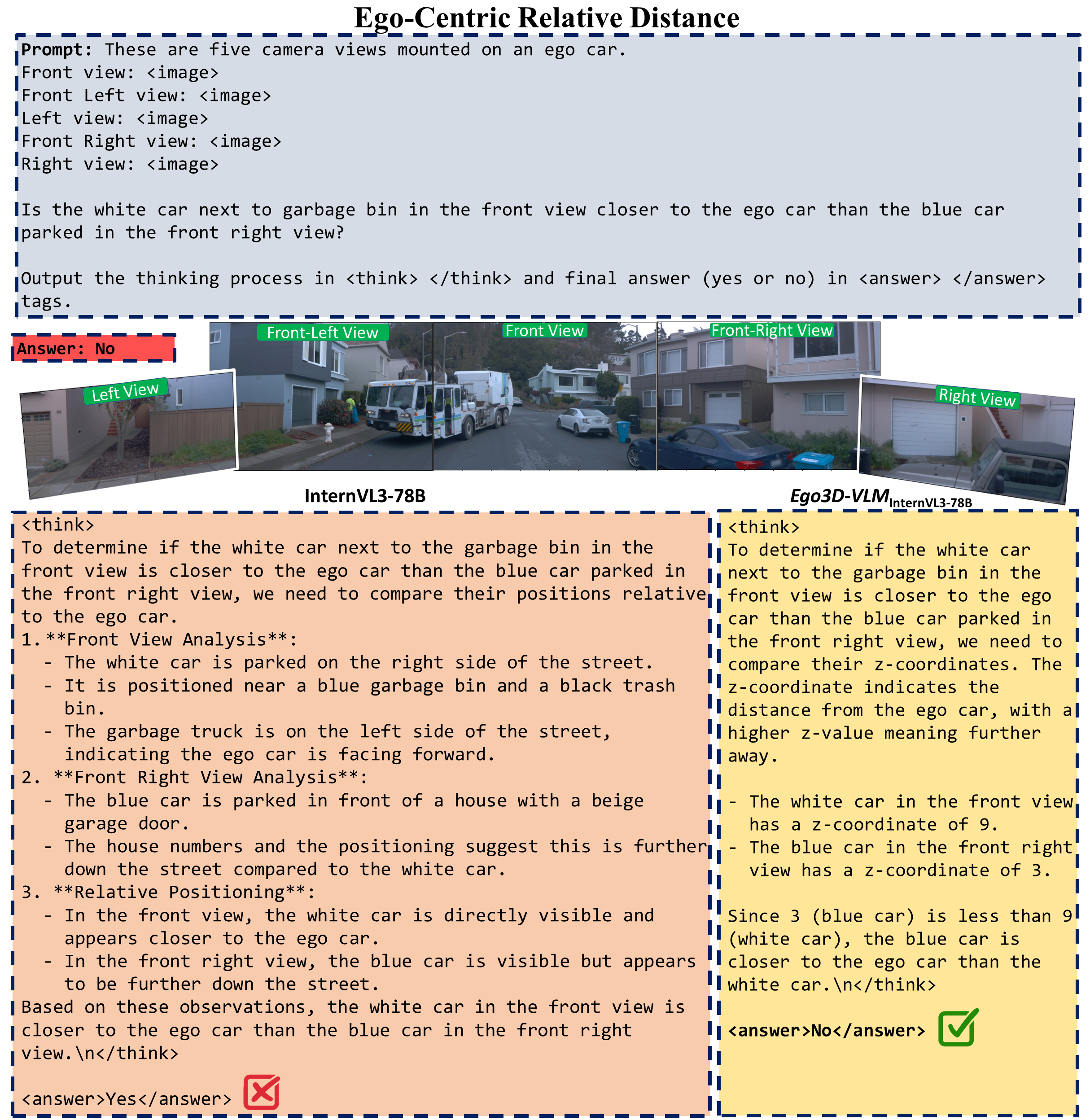}
    \caption{Example responses of the baseline and \textit{Ego3D-VLM} on ego-centric relative distance task.}
    \label{fig:sample10}
\end{figure*}

\begin{figure*}
    \centering    
    \includegraphics[width=1\linewidth]{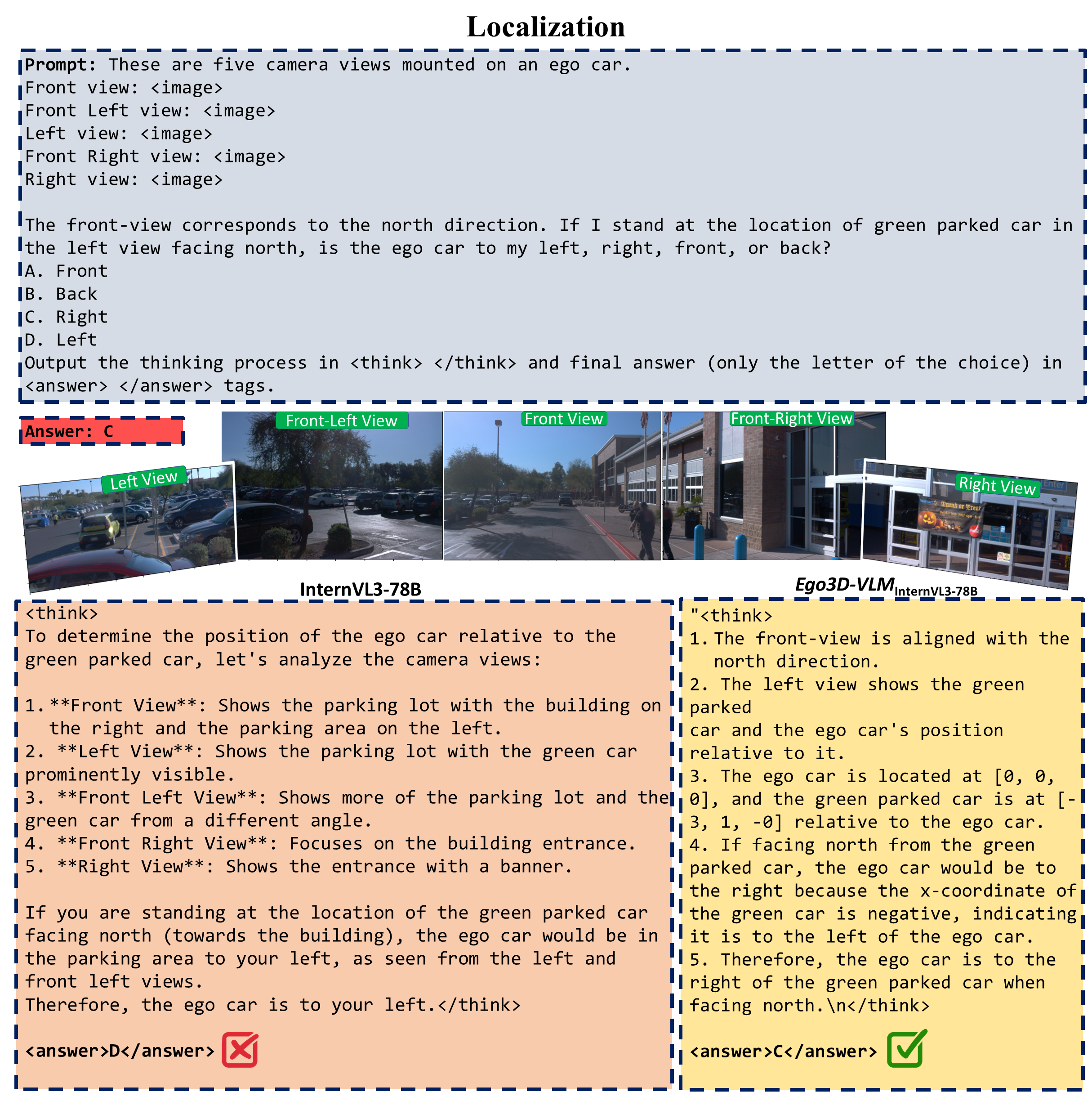}
    \caption{Example responses of the baseline and \textit{Ego3D-VLM} on localization task.}
    \label{fig:sample11}
\end{figure*}

Figures \ref{fig:sample1}-\ref{fig:sample11} demonstrate example responses of InternVL3-78B and \textit{Ego3D-VLM}\textsubscript{InternVL3-78B} on all categories of \texttt{Ego3D-Bench}. As seen, \texttt{Ego3D-Bench} enhances the spatial reasoning ability of the baseline by providing the textual cognitive map.

\subsection{Limitations and Future Work}
\label{ssec:app_limitation}
This work has several limitations that need further investigation. The proposed \textit{Ego3D-VLM} relies on the reasoning capabilities of the underlying vision-language models (VLMs). Consequently, for models with limited reasoning ability (e.g., VILA1.5-8B), we observe little to no improvement when combined with \textit{Ego3D-VLM}.

Additionally, the REC models used in our pipeline provide 2D bounding box locations for all expressions in the prompt. This results in redundant information within the cognitive maps, which may confuse or mislead the VLMs. Our ablation studies demonstrated that when the list of objects mentioned in the prompt is known in advance, the RMSE improves by approximately 1.5 meters.

Another limitation lies in the accuracy of metric depth estimation in outdoor environments. To mitigate this, we proposed a relational scaling approach based on common-sense object sizes. However, this method is inherently approximate and not fully reliable.
Future work should aim to address these issues by enhancing the depth understanding of VLMs or by improving metric depth estimation in outdoor settings. Better integration of spatial reasoning and scene geometry into VLMs could further improve performance in complex 3D environments. Another major limitation in spatial reasoning datasets and benchmarks is the availability of ground-truth 3D. Previous work has shown the effectiveness of neural radiance fields in generating photo-realistic 3D data \cite{posegen}. Future work should investigate this direction for extending spatial reasoning dataset without having ground-truth 3D annotations. 

\subsection{Statement on LLMs Assistance}
We declare that some portions of this document have been lightly refined using Large Language Models (e.g., ChatGPT) to enhance clarity and polish. All substantive content and ideas remain entirely our own.

\clearpage
\clearpage
\clearpage
\clearpage
\clearpage
\clearpage
\clearpage
\clearpage
\clearpage
\clearpage

\end{document}